\documentclass{article} 
\usepackage{iclr2024_conference,times}

\usepackage{amsmath,amsfonts,bm}

\def\eqref#1{equation~\ref{#1}}

\def\1{\bm{1}}

\DeclareMathAlphabet{\mathsfit}{\encodingdefault}{\sfdefault}{m}{sl}
\SetMathAlphabet{\mathsfit}{bold}{\encodingdefault}{\sfdefault}{bx}{n}

\usepackage{hyperref}
\usepackage{url}
\usepackage{graphicx}
\usepackage{svg}
\usepackage{bm}
\usepackage{amsmath}
\usepackage{wrapfig}
\usepackage{amsfonts,amsmath,amssymb,amsthm}
\usepackage{algorithm}
\usepackage{algpseudocode}
\usepackage{booktabs}

\title{Bidirectional Temporal Diffusion Model for Temporally Consistent Human Animation}

\author{
  Tserendorj Adiya$^{3,}$\thanks{Work done during internship at Korea Institute of Science and Technology.}
  \quad Jae Shin Yoon$^2$ \quad Jungeun Lee$^{1}$ \quad Sanghun Kim$^{1}$  \quad Hwasup Lim$^1$
  \\\\ \quad \quad $^1$Korea Institute of  Science and Technology \quad $^2$Adobe \quad $^3$AI Center, CJ Corporation
  \\\\ \quad {\small \texttt{ts.adiya@cj.net}, \texttt{jaeyoon@adobe.com}, \texttt{\{092599,kei97103,hslim\}@kist.re.kr}}
  \vspace{-5mm}
}
\iclrfinalcopy 
\begin{document}

\maketitle

\begin{abstract}
We introduce a method to generate temporally coherent human animation from a single image, a video, or a random noise.
This problem has been formulated as modeling of an auto-regressive generation, \textit{i.e.}, to regress past frames to decode future frames.
However, such unidirectional generation is highly prone to motion drifting over time, generating unrealistic human animation with significant artifacts such as appearance distortion. 
We claim that \textit{bidirectional} temporal modeling enforces temporal coherence on a generative network by largely suppressing the appearance ambiguity.
To prove our claim, we design a novel human animation framework using a denoising diffusion model: 
a neural network learns to generate the image of a person by denoising temporal Gaussian noises whose intermediate results are cross-conditioned bidirectionally between consecutive frames. 
In the experiments, our method demonstrates strong performance compared to existing unidirectional approaches with realistic temporal coherence. \href{https://typest.github.io/btdm/}{Project Page.}

\end{abstract}

\section{Introduction}
\begin{figure*}[t]
    \centering
    \includegraphics[width=1\textwidth]{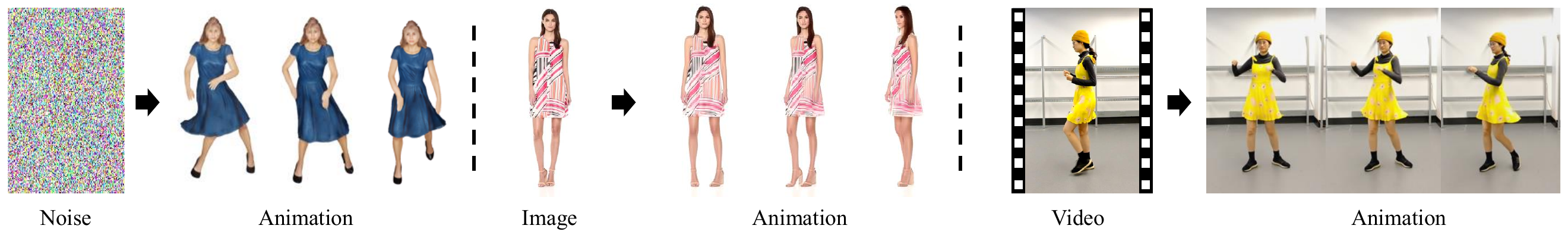}
    \vspace{-5mm}
    \caption{Our method generates temporally coherent human animation from various modalities.}
    \label{fig:teaser}
\end{figure*}

Humans express their own space-time continuum in the form of appearance and motion. While existing generative models~\cite{isola2017image,sarkar2021humangan} have been successful to restore the space, \textit{i.e.}, high-quality image generation with diverse human appearance, they often fail to decode the time, \textit{e.g.}, temporally incoherent human motion. In this paper, we introduce a method for temporal modeling of a generative network to synthesize temporally consistent human animations. Our method can generate a human animation from three different modalities: a random noise, a single image, and a single video as shown in Figure.~\ref{fig:teaser}. Such generated human animations enable a number of applications including novel content creation for non-expert media artists and pre-visualization of human animation that can be further refined by professional video creators.

\begin{wrapfigure}{R}{0.4\textwidth}
\vspace{-7mm}
  \begin{center}
    \includegraphics[width=0.4\textwidth]{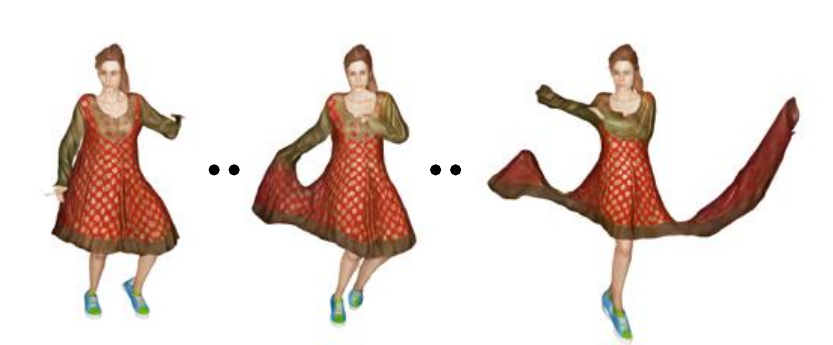}
  \end{center}
    \vspace{-5mm}
  \caption{\small Results from a unidirectional generative model with texture drifting over time.}
  \vspace{-3mm}
  \label{fig:toy_exp}
\end{wrapfigure}

The temporal modeling for human animation has been often formulated as a video auto-regression problem: using past frames as a condition to decode future frames. While such unidirectional generation (forward auto-regression) has shown smooth animation results, it often suffers from texture drifting, \textit{e.g.}, the texture on the clothing of a person such as a skirt in Figure~\ref{fig:toy_exp} is largely distorted along its dynamic movements. This is mainly due to the significant motion-appearance ambiguity where there exist infinite solutions to decide the future state of human appearance even with the same motion, which amplifies the artifacts (e.g., distortion) over time.

To suppress such motion-appearance ambiguity, we model a human appearance bidirectionally: a generative network decodes the human appearance in the context of both forward and backward image regression whose intermediate features are cross-conditioned over time. Our key observation is that the bidirectional temporal consistency in feature space highly suppresses the motion-appearance ambiguity, which prevents from the texture drifting while maintaining its temporal smoothness.

\begin{figure*}
    \centering
    \includegraphics[width=1\textwidth]{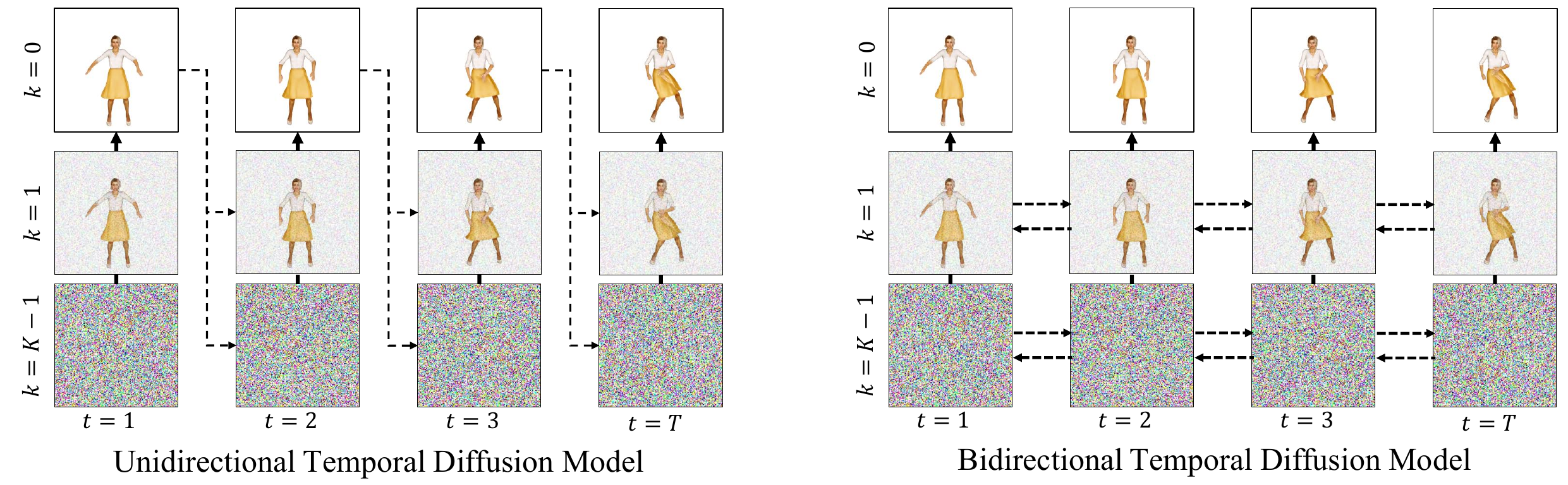}
    \vspace{-4mm}
    \caption{The left illustration represents a unidirectional diffusion model, and the right one provides an overview of our proposed bidirectional temporal diffusion model (BTDM). The dotted arrows indicate the direction of conditioning, and $k$ and $t$ represent the denoising step and time interval, respectively.}
        \vspace{-5mm}
    \label{fig:main}
\end{figure*}

We realize the idea of bidirectional temporal modeling by utilizing a generative denoising diffusion model~\cite{ho2020denoising}. A denoising network learns to iteratively remove temporal Gaussian noises to generate the human animation guided by conditioning poses and appearance style. Inspired by message passing algorithms in dynamic programming~\cite{felzenszwalb2010dynamic,arora2009message}, we recursively cross-condition the intermediate results between consecutive frames in a bidirectional way as shown in Figure~\ref{fig:main}; where the temporal context of human appearance is locally consistent for consecutive frames at the first denoising step, and it is progressively refined at every denoising iteration to be globally coherent for entire frames.

In the experiments, we demonstrate that our bidirectional denoising diffusion model generates human animations from a single image with a strong temporal coherence, outperforming the results from unidirectional generative models. We also show that learning from multiple frames, \textit{i.e.}, a person-specific video, can further improve the physical plausibility of the generated human animation. Finally, we showcase that our method can generate human animations with diverse clothing styles and identities without any conditioning images.

\noindent\textbf{Contribution}
(1) We propose a bidirectional temporal diffusion model that can generate temporally coherent human animation from random noise, a single image, or a video.
(2) Inspired by dynamic message passing algorithms, we introduce the feature cross-conditioning between consecutive frames with recursive sampling, which allows embedding the motion context on the iterative denoising process in a locally and globally consistent way.
(3) We quantitatively and qualitatively demonstrate that our method shows a strong temporal coherence compared to existing unidirectional methods. For an accurate evaluation, we newly create high-quality synthetic data of people in dynamic movements using graphics simulation, which provides ground-truth data, \textit{i.e.,} different people in the perfectly same motion.

\section{Related Works}
\noindent\textbf{Human Motion Transfer} Given a sequence of guiding body poses and the style of human appearance, it aims to generate the human animation that satisfies the conditioning motion and style. Many existing pose transfer methods have utilized 2D keypoints as conditioning body pose maps~\cite{chan2019everybody, balakrishnan2018synthesizing, esser2018variational, liu2019neural}. However, these approaches often fail to extract the physical implications from the keypoints maps, resulting in temporally unnatural human animation.

To address this motion consistency issue, methods such as EDN~\cite{chan2019everybody}, V2V~\cite{wang2018video}, and DIW~\cite{wang2021dance} leveraged Markovian independence to generate auto-regressive frames. These approaches utilize Densepose~\cite{guler2018densepose} as a 2D pose conditioning and learn motion-dependent appearance for a specific person, producing realistic animation results for unseen motions. Recent advancements in this area involve embedding 3D velocities from the SMPL~\cite{loper2015smpl} model as pose conditioning~\cite{yoon2022learning}, leading to the better generation of complex transformations. However, these methods require extensive training on the videos of a single individual, limiting their generalizability to diverse people.

To synthesize human animations of diverse people using a single model, several works have studied human motion transfer from a single image. Solutions include applying affine transformations~\cite{balakrishnan2018synthesizing, zhou2019dance}, flow-based warping~\cite{wang2019few, zhao2022thin, siarohin2021motion, siarohin2019first}, or assuming a base 3D human model and texture mapping with DensePose~\cite{neverova2018dense, huang2021few} or the SMPL model~\cite{li2019dense, liu2019liquid}. However, these methods struggle to represent diverse surface transformations in clothing, \textit{i.e.,} the clothing texture looks static even under the pose changes, resulting in unnatural animations.

\textbf{Generative Diffusion Models} Recently, diffusion models have demonstrated outstanding performance in high-quality image generation~\cite{ho2020denoising, nichol2021improved, song2020score}, text-to-image translation~\cite{rombach2022high, preechakul2022diffusion, saharia2022photorealistic, ramesh2021zero}, image super-resolution~\cite{saharia2022image}, image restoration~\cite{kawar2022denoising, wang2022zero} view synthesis~\cite{watson2022novel}. Compared to generative adversarial networks (GANs)~\cite{isola2017image,karras2019style}, diffusion models enable more stable training and reduced mode collapse, leading to diverse and high-quality generation results.

The initial diffusion model was based on Song's~ \cite{song2019generative} score-matching approach, which estimates gradients using Langevin dynamics to infer data distributions. Subsequently, the DDPM~\cite{ho2020denoising} method was introduced, leveraging weighted variational bounds and becoming widely adopted. Later, NCSN~\cite{song2019generative} and its equivalent from ODE~\cite{song2020score} emerged, presenting a more general form.

One notable drawback of Markov Chain Monte Carlo (MCMC) based inference in diffusion models is the longer inference time compared to GANs. DDIM~\cite{song2020denoising} addresses this issue by interpreting the diffusion process as an implicit function that significantly reducing sampling time while preserving generation quality.

Diffusion models, as referenced in recent studies such as ~\cite{ho2022video, yang2023diffusion, ho2022imagen, singer2022make, zhou2022magicvideo, esser2023structure, guo2023animatediff}, have become increasingly popular in the field of video generation. These models often utilize techniques like cross-attention and 3D U-Nets to ensure videos remain consistent over time. Despite their potential, most of these methods face challenges in generating longer videos without encountering issues like shape drifting and appearance jitters. This paper presents a new and practical approach to overcome these limitations. Our method, which is distinct from previous work, employs a 2D-Unet based framework to create temporally coherent animations. Notably, our approach is effective in generating animations of humans and is not constrained by video length.

\section{Method}
Conventional Denoising Diffusion Probabilistic Models (DDPM) work by gradually diffusing isotropic Gaussian noise onto a data sample $y \in \mathcal{D}$ across $K$ steps along a Markovian chain. The process is reversed, such that $y$ is approximated from the $\mathcal{N}(0,I)$ distribution. 

One can extend these conventional DDPM to generate a human animation driven by a sequence of human pose maps $\mathcal{S} = \{s_1,...,s_T\}$ (\textit{e.g.}, densepose~\cite{guler2018densepose}) in an auto-regressive way. For example, a network is designed to generate future frames dependent on previous frames by gradually diffusing isotropic Gaussian noise onto the training sample $y_t \in Y = \{y_1, ..., y_T\}$ under the conditional Markovian independence, \textit{i.e.}, $p(Y)= \prod_{t=1}^T p(y_t | y_{t-1};s_t\in\mathcal{S})$. However, such autoregressive models often suffer from texture drifting due to the motion-appearance ambiguity that is inherent in unidirectional prediction.

To suppress the motion-appearance ambiguity, we design a bidirectional temporal diffusion model (BTDM) as shown in Figure~\ref{fig:main}. BTDM learns motion-dependent appearances in both forward and backward directions along the time axis. The denoising results from each step in either time direction serve as mutual conditions for generating human animation. Our model can generate realistic animations unconditionally, as well as conditionally from a single image or video.

\subsection{Bidirectional Temporal Diffusion Model}

Given a pose sequence $S$ and its corresponding image sequence $Y$, modeling their mapping bidirectionally along the time axis that follows Markovian independence results in:

\begin{equation}
    p_f(Y|S):=\prod_{t=1}^T p(y_{t}|y_{t-1}, s_t), \quad \quad p_b(Y|S):=\prod_{t=1}^T p(y_{t-1}|y_{t}, s_{t-1})
\label{eq: prob_orig}
\end{equation}

In this setup, $p_f$ represents the forward direction along the time axis, and $p_b$ signifies the backward direction. We define a marginal distribution with isotropic Gaussian process that gradually adds increasing amounts of noise to the data sample as the signal-to-noise-ratio $\lambda(\cdot)$ decreases, following~\cite{salimans2021progressive}:

\begin{equation}
q(y_t^{1:K}|y_t^0) := \prod_{k=1}^K q(y_t^{k}|y_t^{k-1}), \quad \quad q(y_t^{k}|y_t^{k-1}) := \mathcal{N}(y_t^{k}; \sqrt{\sigma(\lambda(k))}y_t^{k-1}, \sigma(-\lambda(k)) \textbf{I})
\label{eq: prob_noise}
\end{equation}

where $\sigma(\cdot)$ is the sigmoid function, $K$ is the number of diffusion step, and $\textbf{I}$ denotes the identity.

Both the motion-dependent appearance distribution in Equation~\ref{eq: prob_orig} and the diffusion process in Equation~\ref{eq: prob_noise} follow a Markovian chain. Ideally, we should predict $y_t$ and $y_{t-1}$ using perfectly denoised $y_{t-1}^0$ (in the forward direction) or $y_{t}^0$ (in the backward direction) as conditions. However, such perfectly denoised images are not available during inference, which leads to overfitting to the training data and amplifies the error from motion-appearance ambiguity. For this reason, we integrate these two independent Markovian chains as follows:


\begin{equation}
    p_f(Y^{k}|S):=\prod_{t=1}^T p(y_{t}^{k}|y_{t-1}^{k}, s_t), \quad \quad p_b(Y^{k}|S):=\prod_{t=1}^T p(y_{t-1}^{k}|y_{t}^{k}, s_{t-1}),
\label{eq: btdm}
\end{equation}

where by utilizing the noisy $y_k$ as a condition, we concurrently diminish the reliance of motion-dependent appearance generation on the preceding frame and avert overfitting to the condition, thereby alleviating artifacts when generating unseen conditions and improving the model's generalization performance. This approach also yields more temporally consistent animations by highly limiting the motion diversity between consecutive frames.

Although $p_f$ and $p_b$ are independent, $p(y_t | y_{t-1})$ and $p(y_{t-1} | y_t)$ are concurrently defined on the time axis $t$. This allows us to optimize both probabilities simultaneously. Therefore, the objective function for training is defined as follows:

\begin{equation}
\begin{split}
    L = \mathop{\mathbb{E}_{t \sim [1,T], k \sim [1,K], y^k \sim q_k, d_f, d_b, c}}
    \Big{[}\frac{1}{2}\big{(}||&f_{\theta}(y_{t}^{k}, y_{t-1}^{k}, \lambda(k), s_{t}, c, d_f) - y_{t}^0||_2^2 \\
    + ||&f_{\theta}(y_{t-1}^{k}, y_{t}^{k}, \lambda(k), s_{t-1}, c, d_b) - y_{t-1}^0||_2^2\big{)}\Big{]}
\end{split}
\label{eq: loss}
\end{equation}

where $f_{\theta}$ is a neural network whose task is to denoise the frame $y_{t-1}^k$, $y_{t}^k$ given a different noisy frame $y_{t}^k$, $y_{t-1}^k$ and given pose $s_t$, $s_{t-1}$. The $\lambda$ is the log signal-to-noise-ratio function dependent on $k$, and $c$ is a single image condition that determines the appearance of a target person. The notation $d_f, d_b$ are learnable positional encoding vectors for distinguishing temporal direction. 
Following the method used in~\cite{ramesh2022hierarchical}, we adapt our model to predict clean images instead of noise. 


\begin{figure*}[t]
    \centering
    \includegraphics[width=1\textwidth]{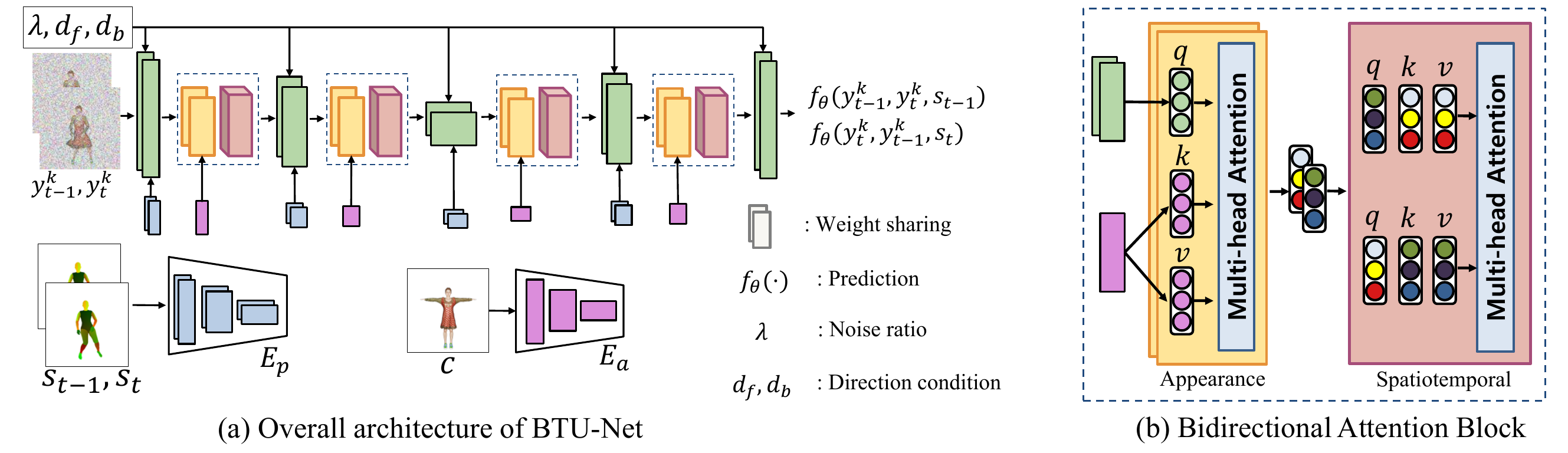}
    \vspace{-4mm}
    \caption{The illustration of (a) our BTU-Net and (b) bidirectional attention block. The dotted squares in (a) represents bidirectional attention block. The small blue and pink squares in (a) indicate the intermediate feature of $E_p$ and $E_a$, respectively.}
    \label{fig:BTUNet}
    \vspace{-4mm}
\end{figure*}
\subsection{Bidirectional Temporal U-Net}
To enable BTDM, we construct a Bidirectional Temporal U-Net (BTU-Net) by modifying the U-Net architecture, as shown in Figure \ref{fig:BTUNet}. This architecture consists of a network, $E_a$, that encodes a single image condition $c$; another network, $E_p$, that encodes poses $s$ corresponding to $t-1$ and $t$; and a pair of U-Nets, $f_{\theta}$, that accept $y_{t-1}^k$ and $y_t^k$ as input and predict the denoised human images temporally in both forward and backward directions. The multi-scale intermediate features are modulated by pose features, noise ratio $\lambda$, and temporal direction vectors ($d_f$ and $d_b$) using existing Feature-wise Linear Modulation layer (FiLM)~\cite{perez2018film}. This pair of U-Nets shares weights and applies attention between the features encoded by $E_a$ and intermediate features of $f_{\theta}$, as shown in the bidirectional attention block in Figure \ref{fig:BTUNet}(b), which is composed of appearance and spatiotemporal block. Each block utilizes multi head attention~\cite{vaswani2017attention}.

\noindent\textbf{Appearance Block} 
The yellow box in Figure \ref{fig:BTUNet}(b), the appearance attention block is composed of parallel multi-head attention mechanisms with weight sharing. For the multi-head attention, features from times \( t \) and \( t-1 \) of \( f_{\theta} \) are used as the query, while the appearance features of \( E_a \) are employed as the key and value. This block is specifically designed to learn the correlation between the appearance features encoded by \( E_a \) and the intermediate features of \( f_{\theta} \). Through the multi-head attention mechanism, it attends to global context correlations, facilitating the effective transfer of appearance features to \( f_{\theta} \) even during rapid actions (those with substantial pose changes relative to frame intervals). This, in turn, enables the generation of novel poses effectively.


\noindent\textbf{Spatiotemporal Block} 
The spatiotemporal attention block takes the output feature pairs from the appearance attention block as inputs and, as illustrated in Figure \ref{fig:BTUNet} (b), performs cross attention. Such a structure effectively enables the learning of the temporal correlations of spatial features between times \( t \) and \( t-1 \). Furthermore, by making the features for generating \( t \) dependent on \( t-1 \) and vice versa, a temporally bidirectional structure is achieved. This design facilitates the efficient learning of temporal correlations.


We adopt the bidirectional attention block for the feature at specific resolutions, \textit{i.e}., $32 \times 32$, $16 \times 16$, and $8 \times 8$. More details on the architecture can be found in Appendix~\ref{appendix:btunet}.

\subsection{Training and Inference for Various Tasks}

\noindent \textbf{Single Image Animation}
 Our BTDM, trained on multiple videos, can be directly applied to generate realistic human animation results for unseen people and poses. Similar to existing one-shot generation methods~\cite{wang2018video}, we further fine-tune our BTU-Net on the given single image to enhance the visual quality. For this, the conditioning image $c$ is set as a single image sequence $Y=\{c\}$ and the pose sequence $S=\{g(c)\}$, where $g(\cdot)$ is a pose estimation function (\textit{e.g}. DensePose). This setup aligns with the training process outlined in Equation \ref{eq: loss}.
 
\noindent\textbf{Person-Specific Animation} Our method can be applied to the task of generating novel animations by training a single person's video. To adapt our method to this task, we train our BTDM framework using the objective function from Equation \ref{eq: loss}, excluding the image condition $c$.

\noindent\textbf{Unconditional animation} 
Moreover, our method facilitates the creation of temporally consistent animations without any appearance-related conditions. For such unconditional generation, we trained our model with the condition $c$ set to $\emptyset$ of Equation \ref{eq: loss}.

During the inference stage, to effectively utilize our BTDM, we employ a bidirectional recursive sampling method across all tasks. More details about this method can be found in Appendix \ref{appendix:BRS}.

\section{Experiments}
We validate our bidirectional temporal diffusion model on two tasks: generating human animation from a single image and generating human animation by learning from a person-specific video. We also show that our model can generate diverse human animation with an unconditional setting (i.e., generating human animation from random noise).

\subsection{Single Image Animation}\label{sec:exp1}

\noindent\textbf{Dataset}
We use two datasets that can effectively validate the quantity and quality of temporal coherence in the generated human animation.
1) Graphics simulation: for quantitative evaluation, we construct a high-quality synthetic dataset using a graphics simulation tool for soft 3D clothing animation~\cite{iClone8} which provides perfect ground truth data for the motion transfer task (\textit{i.e.,} different people in the exact same motion, which does not exist from real-world videos) with physically plausible dynamic clothing movements. The dataset includes a total of 80 training videos and 19 testing videos, each of which lasts 32 seconds at 30 FPS. We customize the 3D human appearance using CharacterCreator~\cite{chacrea}, and we use Mixamo motion data~\cite{mixamo} for animation. The pose map is obtained by rendering the IUV surface coordinates of a 3D body model (\textit{i.e.,} SMPL\cite{loper2015smpl}). Please see the appendix~\ref{appendix:gsdata} for more details of our graphics simulation data.
2) UBC Fashion dataset~\cite{zablotskaia2019dwnet}: it consists of 500 training and 100 testing videos of individuals wearing various outfits and rotating 360 degrees. Each video lasts approximately 12 seconds at 30 FPS. We apply DensePose~\cite{guler2018densepose} to obtain pose UV maps. We use this dataset for the qualitative demonstration on real images since it does not provide ground truth data with the exact same motion.

\noindent\textbf{Baselines}
We compare our method to existing unidirectional temporal models: Thin-Plate Spline Motion Model for Image Animation (TPSMM) \cite{zhao2022thin} and Motion Representations for Articulated Animation (MRAA) \cite{siarohin2021motion} are designed to predict forward optical flow to transport the pixel from a source to target pose, following a rendering network. Both methods were trained on each dataset from scratch using the provided scripts and recommended training setup. All methods are trained at a resolution of 256$\times$256.

\noindent\textbf{Metric} 
To evaluate the quality of the generated human animations, we employ five key metrics: 1) SSIM (Structural Similarity Index)~\cite{wang2004image}: quantifies the structural similarity between the generated and ground truth images based on local patterns of pixel intensities and contrast spaces. 2) LPIPS (Learned Perceptual Image Patch Similarity)~\cite{zhang2018unreasonable}: cognitive similarity between synthesized images and ground truth images by comparing the perceptual features extracted from both, utilizing a pre-trained deep neural network. 3) tLPIPS (Temporal Learned Perceptual Image Patch Similarity)~\cite{chu2020learning}: extends the LPIPS measure to temporal domain, evaluating the plausibility of change across consecutive frames. It is defined as $\mathrm{tLPIPS} = ||\mathrm{LPIPS}(y_t, y_{t-1}) - \mathrm{LPIPS}(g_t, g_{t-1})||$, where $y$ and $g$ represent the synthesized and ground truth images, respectively. 4) tOF~\cite{chu2020learning}: pixel-wise difference of the estimated optical flow between each sequence and the ground truth. 5) FID (Fréchet Inception Distance)~\cite{heusel2017gans}: measures the distance between the distributions of synthesized and real images in the feature space of a pre-trained Inception network.

\begin{figure*}[t]
    \centering
    \includegraphics[width=1\textwidth]{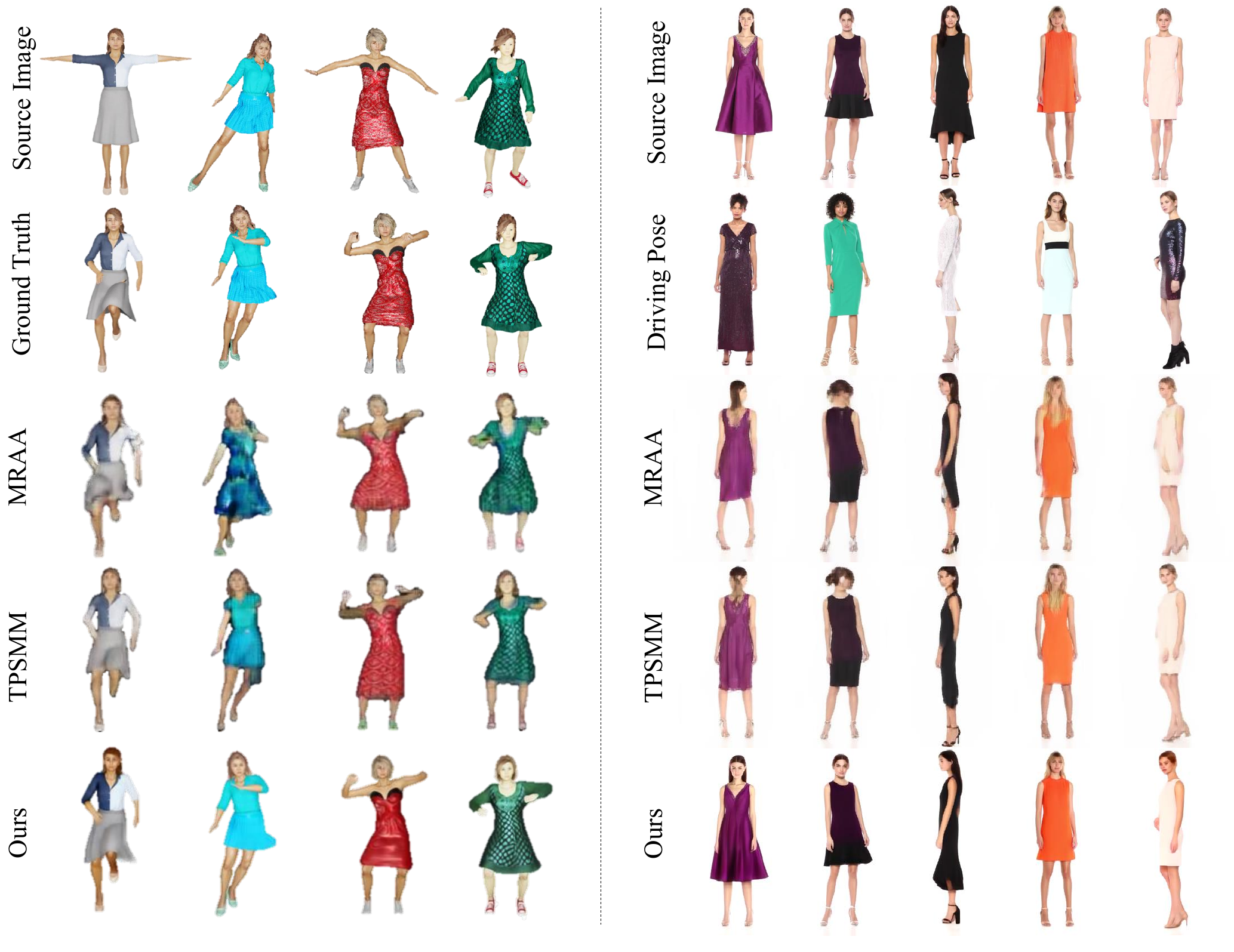}
\vspace{-5mm}
    \caption{Qualitative comparisons for the single image animation task on graphics simulation (left) and UBC Fashion data (right).}
    \label{fig:single_qual}
    \vspace{-4mm}
\end{figure*}

\noindent\textbf{Result} The quantitative results for the graphics simulation data are presented in Table \ref{tab:synth_quant}. Our BTDM method outperforms other methods in all metrics. As can be seen in Figure \ref{fig:single_qual}, our method closely resembles the source image, and the appearance changes depending on the movement are more realistic than other baseline methods. TPSMM and MRAA undergo significant artifacts such as texture distortion and blur due to errors in the forward optical flow prediction. In particular, the models from baseline methods highly confuse on the motion with large dynamics. The same trend is observed in the UBC Fashion data. Specifically, when the appearance of a driven video significantly differs from the source image in TPSMM or MRAA methods, abnormal artifacts often occur such as the loss of identity. Moreover, our method is found to preserve fine details considerably better.

\subsection{Person-specific Animation}
\noindent\textbf{Dataset}
To evaluate the performance of our method in the task of person-specific animation, we use five videos from~\cite{yoon2022learning}. Each video comprises between 6K and 15K frames, featuring a person performing a diverse range of dynamic actions. The pose UV map is obtained using DensePose\cite{guler2018densepose}.

\noindent\textbf{Baseline}
We compare our method to V2V \cite{wang2018video}, EDN \cite{chan2019everybody}, HFMT \cite{kappel2021high}, DIW \cite{wang2021dance}, and MDMT \cite{yoon2022learning}, which utilized a generative network in a temporally unidirectional way. All methods were trained on the training set of each video and evaluated on the test set.

\noindent\textbf{Metrics} We use SSIM, LPIPS, and tLPIPS as used in Section \ref{sec:exp1}.

\begin{table}[t]
  \caption{Quantitative results for single image animation tested on simulation data. }
  \label{tab:synth_quant}
  \centering
  \small
  \begin{tabular}{llllll}
    \toprule
    Methods & SSIM$\uparrow$ & LPIPS$\downarrow$ & tLPIPS$\downarrow$ & tOF  & FID$\downarrow$ \\
    \midrule
    MRAA~\cite{siarohin2021motion}      & 0.894     & 0.140      & 0.011   & 13.58      &   67.68     \\
    TPS~\cite{zhao2022thin}      & 0.915     & 0.077      & 0.005    & 11.92             &   48.76     \\
    Ours     & \textbf{0.958} & \textbf{0.036} & \textbf{0.003}     & \textbf{8.93}      &  \textbf{11.14}  \\
    \bottomrule
  \end{tabular}
  \vspace{-4mm}
\end{table}

\noindent\textbf{Result}
The evaluation results for our method and the baselines on the test sequences from the five videos are displayed in Table \ref{tab:video_quant}. Our approach exhibits a performance that is either comparable to or surpasses that of other state-of-the-art methods in the LPIPS and tLPIPS metrics. Note that, while our model can synthesize the background, we only evaluate the quality of the foreground synthesis for consistent and fair comparison across baseline methods where we use existing segmentation method~\cite{gong2018instance} to remove the background.  for evaluation.  Specifically, our method outshines all others in the SSIM evaluation with a diffusion-based generative framework. The highest average score implies that our method performs consistently better than other methods in terms of temporal coherence and visual plausibility across assorted appearance and motion styles.

\begin{figure*}[t]
    \centering
    \includegraphics[width=0.9\textwidth]{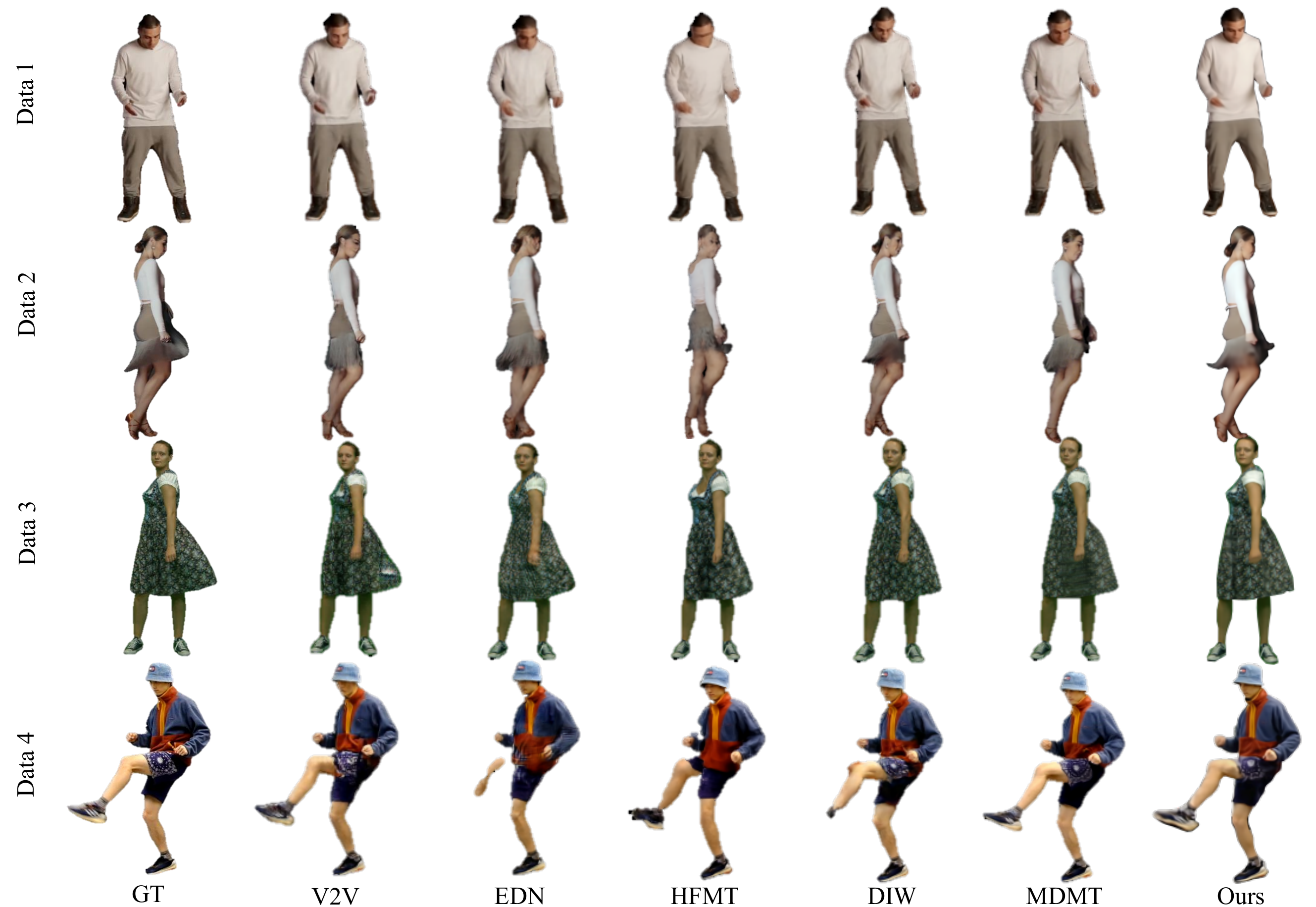}
    \vspace{-3mm}
    \caption{Qualitative comparison between the state-of-the-art methods and our approach for the task of generating person-specific human animation.}
    \label{fig:person_spec}
    \vspace{-3mm}
\end{figure*}

Further qualitative results are demonstrated in Figure \ref{fig:person_spec} where the baseline methods often lose context or become blurred in complex poses, leading to physically implausible human animation. Our method demonstrates robustness to dynamic movements and strong temporal coherence, yielding clear and stable results. Please also refer to the demo video.

\begin{table}
  \caption{Quantitative results of person-specific animation. Each of the three values are in the order of LPIPS($\downarrow$)$\times$10$^2$, tLPIPS($\downarrow$)$\times$10$^3$, and SSIM($\uparrow$)$\times$10, respectively. The number of images used for training is indicated in parentheses, e.g., (6K). }
  \label{tab:video_quant}
  \centering
\hspace*{-1cm}
\small
  \begin{tabular}{cccccc|c}
    \toprule
    Methods & Data 1 (6K) & Data 2 (10K) & Data 3 (10K) & Data 4 (15K) & Data 5 (15K) & Average \\
    \midrule
    V2V    & 1.84/2.95/9.69                     & 3.03/3.83/9.60      & 11.51/3.80/9.05        &   3.06/2.98/9.40   & 4.01/4.04/9.49    & 4.69/3.52/9.45\\
    EDN     & 2.74/3.86/9.57                     & 3.98/5.40/9.46      & 13.12/4.52/8.96       &   4.90/5.09/9.22   & 5.00/4.82/9.34  & 5.95/4.74/9.31  \\
    HFMT     & 3.68/4.41/9.48                     & 6.39/8.54/9.26      & 13.27/4.62/8.91       &   6.08/3.22/9.10   & 6.86/4.53/9.21   & 7.26/5.06/9.19  \\
    DIW     & 1.83/2.88/9.68                     & 2.70/4.11/9.61      & 11.89/4.09/9.03       &   2.83/4.66/9.45   & 4.14/5.20/9.48  & 4.68/4.19/9.45 \\
    MDMT     & \textbf{1.76}/\textbf{2.58}/\textbf{9.73}                    & \textbf{2.68}/3.77/9.65      & 10.48/3.12/9.11       &   2.81/2.86/9.45   & \textbf{3.81}/\textbf{4.12}/9.50  & \textbf{4.31}/3.29/9.49  \\
    Ours     & 1.90/2.83/9.72                    & 2.91/\textbf{3.76}/\textbf{9.65}      & \textbf{10.32}/\textbf{2.52}/\textbf{9.28}       &   \textbf{2.78}/\textbf{2.85}/\textbf{9.57}   & 4.07/4.26/\textbf{9.51}  & 4.39/\textbf{3.22}/\textbf{9.55} \\
    \bottomrule
  \end{tabular}
\end{table}

\begin{table}[t]
  \caption{Quantitative results on ablation study with graphics simulation data. }
  \label{tab:synth_quant_ablation}
  \centering
  \small
  \begin{tabular}{lccc}
    \toprule
    Methods & SSIM$\uparrow$ & LPIPS$\downarrow$ & tLPIPS$\downarrow$ \\
    \midrule
    Ours w/ Unidirectional Model     & 0.937  & 0.052 & 0.005  \\ 
    Ours     & \textbf{0.958} & \textbf{0.036} & \textbf{0.003}   \\
    \bottomrule
  \end{tabular}
\end{table}

\begin{figure*}[t]
    \centering
    \includegraphics[width=0.85\textwidth]{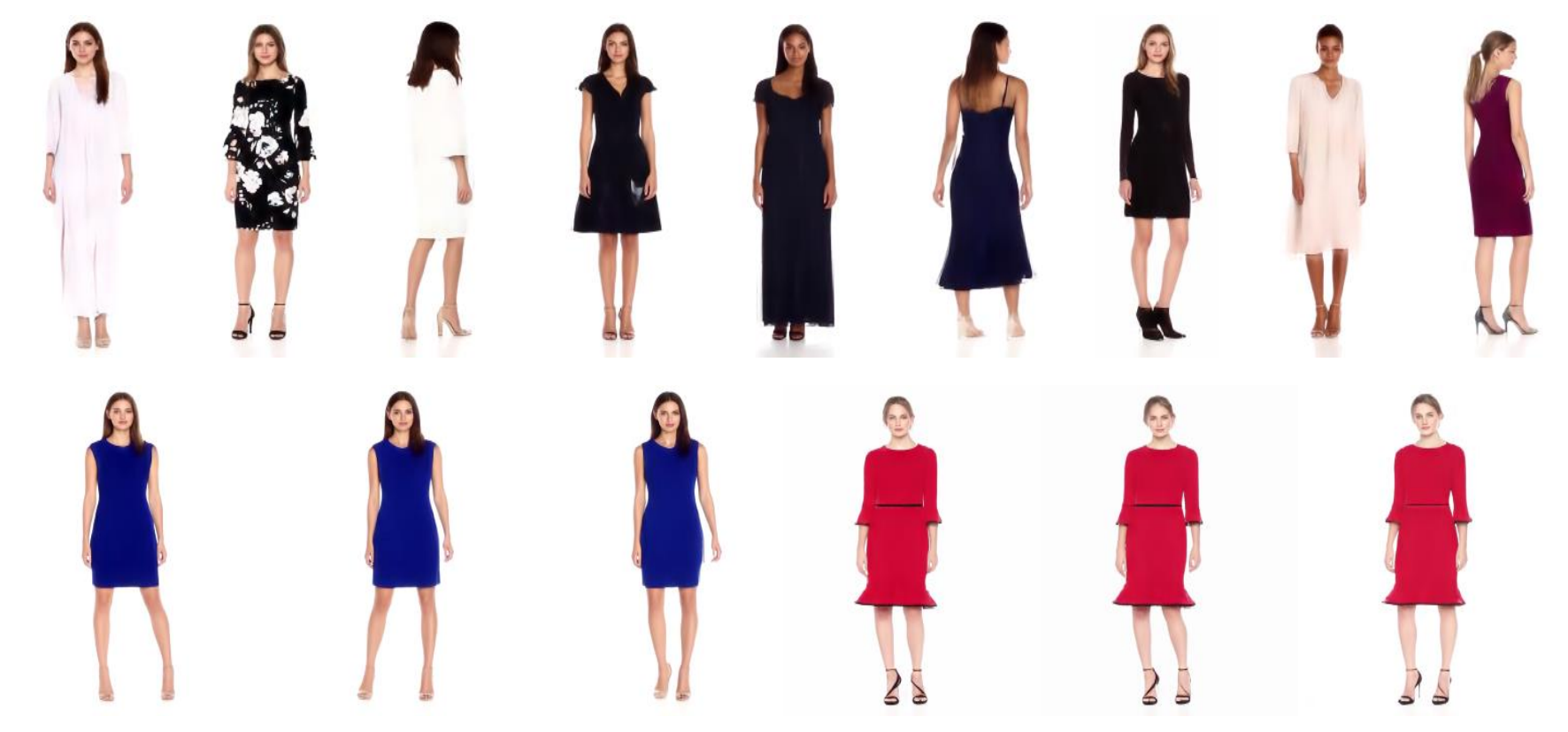}

    \caption{Qualitative results of unconditional animation generation with UBC Fashion data. The top row shows various people generated unconditionally, while the bottom row displays samples of the generated sequences.}
    \label{fig:uncond}
    \vspace{-3mm}
\end{figure*}

\subsection{Ablation study}
To evaluate the effect of the module in our method, we perform an ablation study. The quantitative results are in Table~\ref{tab:synth_quant_ablation}, and please refer to the supplementary materials for visual comparison. 

\noindent\textbf{Unidirectional vs. Bidirectional} We compare our bidirectional temporal diffusion model with the unidirectional one.  For this, we trained the same BTU-Net in a unidirectional manner using the same loss. The bidirectional approach demonstrates far more spatiotemporal consistency based on tLPIPS in Table~\ref{tab:synth_quant_ablation}. Our main observation is that due to significant motion-appearance ambiguity, the generated texture sometimes diverges at the end of the frame under highly dynamic human movements. Based on the improvements in LPIPS, we can notice such strong temporal coherence helps with improving the visual quality as well.


\noindent\textbf{Number of Images for Fine-Tuning} For single-image animation, we fine-tune the model on a single image. We found that the quality improves as the number of fine-tuning images increases. In the supplementary material, we introduce the experiments about the impact of the number of fine-tuning images on rendering quality.

\textbf{Unconditional Human Animation Generation}
Figure~\ref{fig:uncond}, we demonstrate that our method can generate human animation with diverse appearances without conditioning any images or videos.

\section{Conclusion}
We introduce a new method to synthesize temporally coherent human animation from a single image, a video, or a random noise.
We address the core challenge of temporal incoherence from existing generative networks that decode future frames in an auto-regressive way. 
We argue that such unidirectional temporal modeling of a generative network involves a significant amount of motion-appearance ambiguity, leading to the artifacts such as texture drifting.
We suppress the motion-appearance ambiguity by newly designing a bidirectional temporal diffusion model (BTDM): a denoising network progressively removes temporal Gaussian noises whose intermediate results are cross-conditioned over consecutive frames, which allows conditioning locally and globally coherent motion context on our video generation framework.
We perform the evaluation on two different tasks, i.e., human animation from a single image and person-specific human animation, and demonstrate that BTDM shows strong temporal coherence, which also helps to improve the visual quality, compared to existing methods.

\noindent\textbf{Limitation} While BTDM produces temporally coherent human animations, there exist several limitations. Since our model generates the video as a function of the estimated body poses, the errors in the pose estimation affect the rendering quality, \textit{e.g.,} the misdetection of hands produces some appearance distortion around the hand. Due to the inherent ambiguity of 2D pose representation, our method sometimes shows weakness in the sequence with 3D human rotations. Our potential future work is to improve 3D awareness and completeness by utilizing a complete 3D body model, e.g., SMPL~\cite{loper2015smpl}, in our bidirectional temporal diffusion framework.

\section{Acknowledgement}
This work was supported by Institute of Information communications Technology Planning Evaluation (IITP) grant funded by the Korea government(MSIT) (No.RS2023-00225630, Development of Artificial Intelligence for Text-based 3D Movie Generation).

\bibliography{iclr2024_conference}

\begin{thebibliography}{59}
\providecommand{\natexlab}[1]{#1}
\providecommand{\url}[1]{\texttt{#1}}
\expandafter\ifx\csname urlstyle\endcsname\relax
  \providecommand{\doi}[1]{doi: #1}\else
  \providecommand{\doi}{doi: \begingroup \urlstyle{rm}\Url}\fi

\bibitem[Adobe()]{mixamo}
Adobe.
\newblock Adobe mixamo.
\newblock \url{https://www.mixamo.com/#/}.

\bibitem[Arora et~al.(2009)Arora, Daskalakis, and Steurer]{arora2009message}
Sanjeev Arora, Constantinos Daskalakis, and David Steurer.
\newblock Message passing algorithms and improved lp decoding.
\newblock In \emph{Proceedings of the forty-first annual ACM symposium on Theory of computing}, pp.\  3--12, 2009.

\bibitem[Balakrishnan et~al.(2018)Balakrishnan, Zhao, Dalca, Durand, and Guttag]{balakrishnan2018synthesizing}
Guha Balakrishnan, Amy Zhao, Adrian~V Dalca, Fredo Durand, and John Guttag.
\newblock Synthesizing images of humans in unseen poses.
\newblock In \emph{Proceedings of the IEEE conference on computer vision and pattern recognition}, pp.\  8340--8348, 2018.

\bibitem[Chan et~al.(2019)Chan, Ginosar, Zhou, and Efros]{chan2019everybody}
Caroline Chan, Shiry Ginosar, Tinghui Zhou, and Alexei~A Efros.
\newblock Everybody dance now.
\newblock In \emph{Proceedings of the IEEE/CVF international conference on computer vision}, pp.\  5933--5942, 2019.

\bibitem[Chu et~al.(2020)Chu, Xie, Mayer, Leal-Taix{\'e}, and Thuerey]{chu2020learning}
Mengyu Chu, You Xie, Jonas Mayer, Laura Leal-Taix{\'e}, and Nils Thuerey.
\newblock Learning temporal coherence via self-supervision for gan-based video generation.
\newblock \emph{ACM Transactions on Graphics (TOG)}, 39\penalty0 (4):\penalty0 75--1, 2020.

\bibitem[Esser et~al.(2018)Esser, Sutter, and Ommer]{esser2018variational}
Patrick Esser, Ekaterina Sutter, and Bj{\"o}rn Ommer.
\newblock A variational u-net for conditional appearance and shape generation.
\newblock In \emph{Proceedings of the IEEE conference on computer vision and pattern recognition}, pp.\  8857--8866, 2018.

\bibitem[Esser et~al.(2023)Esser, Chiu, Atighehchian, Granskog, and Germanidis]{esser2023structure}
Patrick Esser, Johnathan Chiu, Parmida Atighehchian, Jonathan Granskog, and Anastasis Germanidis.
\newblock Structure and content-guided video synthesis with diffusion models.
\newblock In \emph{Proceedings of the IEEE/CVF International Conference on Computer Vision}, pp.\  7346--7356, 2023.

\bibitem[Felzenszwalb \& Zabih(2010)Felzenszwalb and Zabih]{felzenszwalb2010dynamic}
Pedro~F Felzenszwalb and Ramin Zabih.
\newblock Dynamic programming and graph algorithms in computer vision.
\newblock \emph{IEEE transactions on pattern analysis and machine intelligence}, 33\penalty0 (4):\penalty0 721--740, 2010.

\bibitem[Gong et~al.(2018)Gong, Liang, Li, Chen, Yang, and Lin]{gong2018instance}
Ke~Gong, Xiaodan Liang, Yicheng Li, Yimin Chen, Ming Yang, and Liang Lin.
\newblock Instance-level human parsing via part grouping network.
\newblock In \emph{Proceedings of the European conference on computer vision (ECCV)}, pp.\  770--785, 2018.

\bibitem[G{\"u}ler et~al.(2018)G{\"u}ler, Neverova, and Kokkinos]{guler2018densepose}
R{\i}za~Alp G{\"u}ler, Natalia Neverova, and Iasonas Kokkinos.
\newblock Densepose: Dense human pose estimation in the wild.
\newblock In \emph{Proceedings of the IEEE conference on computer vision and pattern recognition}, pp.\  7297--7306, 2018.

\bibitem[Guo et~al.(2023)Guo, Yang, Rao, Wang, Qiao, Lin, and Dai]{guo2023animatediff}
Yuwei Guo, Ceyuan Yang, Anyi Rao, Yaohui Wang, Yu~Qiao, Dahua Lin, and Bo~Dai.
\newblock Animatediff: Animate your personalized text-to-image diffusion models without specific tuning.
\newblock \emph{arXiv preprint arXiv:2307.04725}, 2023.

\bibitem[Heusel et~al.(2017)Heusel, Ramsauer, Unterthiner, Nessler, and Hochreiter]{heusel2017gans}
Martin Heusel, Hubert Ramsauer, Thomas Unterthiner, Bernhard Nessler, and Sepp Hochreiter.
\newblock Gans trained by a two time-scale update rule converge to a local nash equilibrium.
\newblock \emph{Advances in neural information processing systems}, 30, 2017.

\bibitem[Ho et~al.(2020)Ho, Jain, and Abbeel]{ho2020denoising}
Jonathan Ho, Ajay Jain, and Pieter Abbeel.
\newblock Denoising diffusion probabilistic models.
\newblock \emph{Advances in Neural Information Processing Systems}, 33:\penalty0 6840--6851, 2020.

\bibitem[Ho et~al.(2022{\natexlab{a}})Ho, Chan, Saharia, Whang, Gao, Gritsenko, Kingma, Poole, Norouzi, Fleet, et~al.]{ho2022imagen}
Jonathan Ho, William Chan, Chitwan Saharia, Jay Whang, Ruiqi Gao, Alexey Gritsenko, Diederik~P Kingma, Ben Poole, Mohammad Norouzi, David~J Fleet, et~al.
\newblock Imagen video: High definition video generation with diffusion models.
\newblock \emph{arXiv preprint arXiv:2210.02303}, 2022{\natexlab{a}}.

\bibitem[Ho et~al.(2022{\natexlab{b}})Ho, Salimans, Gritsenko, Chan, Norouzi, and Fleet]{ho2022video}
Jonathan Ho, Tim Salimans, Alexey Gritsenko, William Chan, Mohammad Norouzi, and David~J Fleet.
\newblock Video diffusion models.
\newblock \emph{arXiv preprint arXiv:2204.03458}, 2022{\natexlab{b}}.

\bibitem[Huang et~al.(2021)Huang, Han, Xu, and Zhang]{huang2021few}
Zhichao Huang, Xintong Han, Jia Xu, and Tong Zhang.
\newblock Few-shot human motion transfer by personalized geometry and texture modeling.
\newblock In \emph{Proceedings of the IEEE/CVF Conference on Computer Vision and Pattern Recognition}, pp.\  2297--2306, 2021.

\bibitem[Isola et~al.(2017)Isola, Zhu, Zhou, and Efros]{isola2017image}
Phillip Isola, Jun-Yan Zhu, Tinghui Zhou, and Alexei~A Efros.
\newblock Image-to-image translation with conditional adversarial networks.
\newblock In \emph{Proceedings of the IEEE conference on computer vision and pattern recognition}, pp.\  1125--1134, 2017.

\bibitem[Kappel et~al.(2021)Kappel, Golyanik, Elgharib, Henningson, Seidel, Castillo, Theobalt, and Magnor]{kappel2021high}
Moritz Kappel, Vladislav Golyanik, Mohamed Elgharib, Jann-Ole Henningson, Hans-Peter Seidel, Susana Castillo, Christian Theobalt, and Marcus Magnor.
\newblock High-fidelity neural human motion transfer from monocular video.
\newblock In \emph{Proceedings of the IEEE/CVF conference on computer vision and pattern recognition}, pp.\  1541--1550, 2021.

\bibitem[Karras et~al.(2019)Karras, Laine, and Aila]{karras2019style}
Tero Karras, Samuli Laine, and Timo Aila.
\newblock A style-based generator architecture for generative adversarial networks.
\newblock In \emph{Proceedings of the IEEE/CVF conference on computer vision and pattern recognition}, pp.\  4401--4410, 2019.

\bibitem[Kawar et~al.(2022)Kawar, Elad, Ermon, and Song]{kawar2022denoising}
Bahjat Kawar, Michael Elad, Stefano Ermon, and Jiaming Song.
\newblock Denoising diffusion restoration models.
\newblock \emph{Advances in Neural Information Processing Systems}, 35:\penalty0 23593--23606, 2022.

\bibitem[Li et~al.(2019)Li, Huang, and Loy]{li2019dense}
Yining Li, Chen Huang, and Chen~Change Loy.
\newblock Dense intrinsic appearance flow for human pose transfer.
\newblock In \emph{Proceedings of the IEEE/CVF Conference on Computer Vision and Pattern Recognition}, pp.\  3693--3702, 2019.

\bibitem[Liu et~al.(2019{\natexlab{a}})Liu, Xu, Zollhoefer, Kim, Bernard, Habermann, Wang, and Theobalt]{liu2019neural}
Lingjie Liu, Weipeng Xu, Michael Zollhoefer, Hyeongwoo Kim, Florian Bernard, Marc Habermann, Wenping Wang, and Christian Theobalt.
\newblock Neural rendering and reenactment of human actor videos.
\newblock \emph{ACM Transactions on Graphics (TOG)}, 38\penalty0 (5):\penalty0 1--14, 2019{\natexlab{a}}.

\bibitem[Liu et~al.(2019{\natexlab{b}})Liu, Piao, Min, Luo, Ma, and Gao]{liu2019liquid}
Wen Liu, Zhixin Piao, Jie Min, Wenhan Luo, Lin Ma, and Shenghua Gao.
\newblock Liquid warping gan: A unified framework for human motion imitation, appearance transfer and novel view synthesis.
\newblock In \emph{Proceedings of the IEEE/CVF International Conference on Computer Vision}, pp.\  5904--5913, 2019{\natexlab{b}}.

\bibitem[Loper et~al.(2015)Loper, Mahmood, Romero, Pons-Moll, and Black]{loper2015smpl}
Matthew Loper, Naureen Mahmood, Javier Romero, Gerard Pons-Moll, and Michael~J Black.
\newblock Smpl: A skinned multi-person linear model.
\newblock \emph{ACM transactions on graphics (TOG)}, 34\penalty0 (6):\penalty0 1--16, 2015.

\bibitem[Neverova et~al.(2018)Neverova, Guler, and Kokkinos]{neverova2018dense}
Natalia Neverova, Riza~Alp Guler, and Iasonas Kokkinos.
\newblock Dense pose transfer.
\newblock In \emph{Proceedings of the European conference on computer vision (ECCV)}, pp.\  123--138, 2018.

\bibitem[Nichol \& Dhariwal(2021)Nichol and Dhariwal]{nichol2021improved}
Alexander~Quinn Nichol and Prafulla Dhariwal.
\newblock Improved denoising diffusion probabilistic models.
\newblock In \emph{International Conference on Machine Learning}, pp.\  8162--8171. PMLR, 2021.

\bibitem[Nvidia()]{omniverse}
Nvidia.
\newblock Nvidia omniverse.
\newblock \url{https://www.nvidia.com/en-us/omniverse/}.

\bibitem[Perez et~al.(2018)Perez, Strub, De~Vries, Dumoulin, and Courville]{perez2018film}
Ethan Perez, Florian Strub, Harm De~Vries, Vincent Dumoulin, and Aaron Courville.
\newblock Film: Visual reasoning with a general conditioning layer.
\newblock In \emph{Proceedings of the AAAI Conference on Artificial Intelligence}, volume~32, 2018.

\bibitem[Pictures()]{alembic}
Sony Pictures.
\newblock Lucasfilm and sony pictures imageworks release alembic 1.0.
\newblock Sony Pictures Imageworks, Lucasfilm (August 9, 2011).

\bibitem[Preechakul et~al.(2022)Preechakul, Chatthee, Wizadwongsa, and Suwajanakorn]{preechakul2022diffusion}
Konpat Preechakul, Nattanat Chatthee, Suttisak Wizadwongsa, and Supasorn Suwajanakorn.
\newblock Diffusion autoencoders: Toward a meaningful and decodable representation.
\newblock In \emph{Proceedings of the IEEE/CVF Conference on Computer Vision and Pattern Recognition}, pp.\  10619--10629, 2022.

\bibitem[Ramesh et~al.(2021)Ramesh, Pavlov, Goh, Gray, Voss, Radford, Chen, and Sutskever]{ramesh2021zero}
Aditya Ramesh, Mikhail Pavlov, Gabriel Goh, Scott Gray, Chelsea Voss, Alec Radford, Mark Chen, and Ilya Sutskever.
\newblock Zero-shot text-to-image generation.
\newblock In \emph{International Conference on Machine Learning}, pp.\  8821--8831. PMLR, 2021.

\bibitem[Ramesh et~al.(2022)Ramesh, Dhariwal, Nichol, Chu, and Chen]{ramesh2022hierarchical}
Aditya Ramesh, Prafulla Dhariwal, Alex Nichol, Casey Chu, and Mark Chen.
\newblock Hierarchical text-conditional image generation with clip latents.
\newblock \emph{arXiv preprint arXiv:2204.06125}, 2022.

\bibitem[Reallusion({\natexlab{a}})]{chacrea}
Reallusion.
\newblock Character creator.
\newblock \url{https://www.reallusion.com/character-creator/}, {\natexlab{a}}.

\bibitem[Reallusion({\natexlab{b}})]{iClone8}
Reallusion.
\newblock iclone8.
\newblock \url{https://www.reallusion.com/iclone/}, {\natexlab{b}}.

\bibitem[Rombach et~al.(2022)Rombach, Blattmann, Lorenz, Esser, and Ommer]{rombach2022high}
Robin Rombach, Andreas Blattmann, Dominik Lorenz, Patrick Esser, and Bj{\"o}rn Ommer.
\newblock High-resolution image synthesis with latent diffusion models.
\newblock In \emph{Proceedings of the IEEE/CVF Conference on Computer Vision and Pattern Recognition}, pp.\  10684--10695, 2022.

\bibitem[Saharia et~al.(2022{\natexlab{a}})Saharia, Chan, Saxena, Li, Whang, Denton, Ghasemipour, Gontijo~Lopes, Karagol~Ayan, Salimans, et~al.]{saharia2022photorealistic}
Chitwan Saharia, William Chan, Saurabh Saxena, Lala Li, Jay Whang, Emily~L Denton, Kamyar Ghasemipour, Raphael Gontijo~Lopes, Burcu Karagol~Ayan, Tim Salimans, et~al.
\newblock Photorealistic text-to-image diffusion models with deep language understanding.
\newblock \emph{Advances in Neural Information Processing Systems}, 35:\penalty0 36479--36494, 2022{\natexlab{a}}.

\bibitem[Saharia et~al.(2022{\natexlab{b}})Saharia, Ho, Chan, Salimans, Fleet, and Norouzi]{saharia2022image}
Chitwan Saharia, Jonathan Ho, William Chan, Tim Salimans, David~J Fleet, and Mohammad Norouzi.
\newblock Image super-resolution via iterative refinement.
\newblock \emph{IEEE Transactions on Pattern Analysis and Machine Intelligence}, 2022{\natexlab{b}}.

\bibitem[Salimans \& Ho(2021)Salimans and Ho]{salimans2021progressive}
Tim Salimans and Jonathan Ho.
\newblock Progressive distillation for fast sampling of diffusion models.
\newblock In \emph{International Conference on Learning Representations}, 2021.

\bibitem[Sarkar et~al.(2021)Sarkar, Liu, Golyanik, and Theobalt]{sarkar2021humangan}
Kripasindhu Sarkar, Lingjie Liu, Vladislav Golyanik, and Christian Theobalt.
\newblock Humangan: A generative model of human images.
\newblock In \emph{2021 International Conference on 3D Vision (3DV)}, pp.\  258--267. IEEE, 2021.

\bibitem[Siarohin et~al.(2019)Siarohin, Lathuili{\`e}re, Tulyakov, Ricci, and Sebe]{siarohin2019first}
Aliaksandr Siarohin, St{\'e}phane Lathuili{\`e}re, Sergey Tulyakov, Elisa Ricci, and Nicu Sebe.
\newblock First order motion model for image animation.
\newblock \emph{Advances in Neural Information Processing Systems}, 32, 2019.

\bibitem[Siarohin et~al.(2021)Siarohin, Woodford, Ren, Chai, and Tulyakov]{siarohin2021motion}
Aliaksandr Siarohin, Oliver~J Woodford, Jian Ren, Menglei Chai, and Sergey Tulyakov.
\newblock Motion representations for articulated animation.
\newblock In \emph{Proceedings of the IEEE/CVF Conference on Computer Vision and Pattern Recognition}, pp.\  13653--13662, 2021.

\bibitem[Singer et~al.(2022)Singer, Polyak, Hayes, Yin, An, Zhang, Hu, Yang, Ashual, Gafni, et~al.]{singer2022make}
Uriel Singer, Adam Polyak, Thomas Hayes, Xi~Yin, Jie An, Songyang Zhang, Qiyuan Hu, Harry Yang, Oron Ashual, Oran Gafni, et~al.
\newblock Make-a-video: Text-to-video generation without text-video data.
\newblock \emph{arXiv preprint arXiv:2209.14792}, 2022.

\bibitem[Song et~al.(2020{\natexlab{a}})Song, Meng, and Ermon]{song2020denoising}
Jiaming Song, Chenlin Meng, and Stefano Ermon.
\newblock Denoising diffusion implicit models.
\newblock In \emph{International Conference on Learning Representations}, 2020{\natexlab{a}}.

\bibitem[Song \& Ermon(2019)Song and Ermon]{song2019generative}
Yang Song and Stefano Ermon.
\newblock Generative modeling by estimating gradients of the data distribution.
\newblock \emph{Advances in neural information processing systems}, 32, 2019.

\bibitem[Song et~al.(2020{\natexlab{b}})Song, Sohl-Dickstein, Kingma, Kumar, Ermon, and Poole]{song2020score}
Yang Song, Jascha Sohl-Dickstein, Diederik~P Kingma, Abhishek Kumar, Stefano Ermon, and Ben Poole.
\newblock Score-based generative modeling through stochastic differential equations.
\newblock In \emph{International Conference on Learning Representations}, 2020{\natexlab{b}}.

\bibitem[Vaswani et~al.(2017)Vaswani, Shazeer, Parmar, Uszkoreit, Jones, Gomez, Kaiser, and Polosukhin]{vaswani2017attention}
Ashish Vaswani, Noam Shazeer, Niki Parmar, Jakob Uszkoreit, Llion Jones, Aidan~N Gomez, {\L}ukasz Kaiser, and Illia Polosukhin.
\newblock Attention is all you need.
\newblock \emph{Advances in neural information processing systems}, 30, 2017.

\bibitem[Wang et~al.(2018)Wang, Liu, Zhu, Liu, Tao, Kautz, and Catanzaro]{wang2018video}
Ting-Chun Wang, Ming-Yu Liu, Jun-Yan Zhu, Guilin Liu, Andrew Tao, Jan Kautz, and Bryan Catanzaro.
\newblock Video-to-video synthesis.
\newblock \emph{arXiv preprint arXiv:1808.06601}, 2018.

\bibitem[Wang et~al.(2019)Wang, Liu, Tao, Liu, Catanzaro, and Kautz]{wang2019few}
Ting-Chun Wang, Ming-Yu Liu, Andrew Tao, Guilin Liu, Bryan Catanzaro, and Jan Kautz.
\newblock Few-shot video-to-video synthesis.
\newblock \emph{Advances in Neural Information Processing Systems}, 32, 2019.

\bibitem[Wang et~al.(2021)Wang, Ceylan, Singh, and Mitra]{wang2021dance}
Tuanfeng~Y Wang, Duygu Ceylan, Krishna~Kumar Singh, and Niloy~J Mitra.
\newblock Dance in the wild: Monocular human animation with neural dynamic appearance synthesis.
\newblock In \emph{2021 International Conference on 3D Vision (3DV)}, pp.\  268--277. IEEE, 2021.

\bibitem[Wang et~al.(2022)Wang, Yu, and Zhang]{wang2022zero}
Yinhuai Wang, Jiwen Yu, and Jian Zhang.
\newblock Zero-shot image restoration using denoising diffusion null-space model.
\newblock In \emph{The Eleventh International Conference on Learning Representations}, 2022.

\bibitem[Wang et~al.(2004)Wang, Bovik, Sheikh, and Simoncelli]{wang2004image}
Zhou Wang, Alan~C Bovik, Hamid~R Sheikh, and Eero~P Simoncelli.
\newblock Image quality assessment: from error visibility to structural similarity.
\newblock \emph{IEEE transactions on image processing}, 13\penalty0 (4):\penalty0 600--612, 2004.

\bibitem[Watson et~al.(2022)Watson, Chan, Brualla, Ho, Tagliasacchi, and Norouzi]{watson2022novel}
Daniel Watson, William Chan, Ricardo~Martin Brualla, Jonathan Ho, Andrea Tagliasacchi, and Mohammad Norouzi.
\newblock Novel view synthesis with diffusion models.
\newblock In \emph{The Eleventh International Conference on Learning Representations}, 2022.

\bibitem[Yang et~al.(2023)Yang, Srivastava, and Mandt]{yang2023diffusion}
Ruihan Yang, Prakhar Srivastava, and Stephan Mandt.
\newblock Diffusion probabilistic modeling for video generation.
\newblock \emph{Entropy}, 25\penalty0 (10):\penalty0 1469, 2023.

\bibitem[Yoon et~al.(2022)Yoon, Ceylan, Wang, Lu, Yang, Shu, and Park]{yoon2022learning}
Jae~Shin Yoon, Duygu Ceylan, Tuanfeng~Y Wang, Jingwan Lu, Jimei Yang, Zhixin Shu, and Hyun~Soo Park.
\newblock Learning motion-dependent appearance for high-fidelity rendering of dynamic humans from a single camera.
\newblock In \emph{Proceedings of the IEEE/CVF Conference on Computer Vision and Pattern Recognition}, pp.\  3407--3417, 2022.

\bibitem[Zablotskaia et~al.(2019)Zablotskaia, Siarohin, Zhao, and Sigal]{zablotskaia2019dwnet}
Polina Zablotskaia, Aliaksandr Siarohin, Bo~Zhao, and Leonid Sigal.
\newblock Dwnet: Dense warp-based network for pose-guided human video generation.
\newblock \emph{arXiv preprint arXiv:1910.09139}, 2019.

\bibitem[Zhang et~al.(2018)Zhang, Isola, Efros, Shechtman, and Wang]{zhang2018unreasonable}
Richard Zhang, Phillip Isola, Alexei~A Efros, Eli Shechtman, and Oliver Wang.
\newblock The unreasonable effectiveness of deep features as a perceptual metric.
\newblock In \emph{Proceedings of the IEEE conference on computer vision and pattern recognition}, pp.\  586--595, 2018.

\bibitem[Zhao \& Zhang(2022)Zhao and Zhang]{zhao2022thin}
Jian Zhao and Hui Zhang.
\newblock Thin-plate spline motion model for image animation.
\newblock In \emph{Proceedings of the IEEE/CVF Conference on Computer Vision and Pattern Recognition}, pp.\  3657--3666, 2022.

\bibitem[Zhou et~al.(2022)Zhou, Wang, Yan, Lv, Zhu, and Feng]{zhou2022magicvideo}
Daquan Zhou, Weimin Wang, Hanshu Yan, Weiwei Lv, Yizhe Zhu, and Jiashi Feng.
\newblock Magicvideo: Efficient video generation with latent diffusion models.
\newblock \emph{arXiv preprint arXiv:2211.11018}, 2022.

\bibitem[Zhou et~al.(2019)Zhou, Wang, Fang, Bui, and Berg]{zhou2019dance}
Yipin Zhou, Zhaowen Wang, Chen Fang, Trung Bui, and Tamara Berg.
\newblock Dance dance generation: Motion transfer for internet videos.
\newblock In \emph{Proceedings of the IEEE/CVF International Conference on Computer Vision Workshops}, pp.\  0--0, 2019.

\end{thebibliography}
\bibliographystyle{iclr2024_conference}

\appendix
\section{Bidirectional Recursive Sampling}
\label{appendix:BRS}
To effectively utilize our Bidirectional Temporal Diffusion Model (BTDM) during the inference stage, we have employed a bidirectional recursive sampling method, which proceeds as follows.

\begin{algorithm}
    \caption{Bidirectional Recursive Sampling}
    \begin{algorithmic}
        \State \textbf{Input:} Initial noisy inputs $Y^k=\{y_1^{k}, ..., y_t^{k}\}$, driven pose sequence $\mathcal{S} = \{s_1,...,s_T\}$.
        \State \textbf{Output:} Denoised animation $Y^0=\{y_1^{0}, ..., y_t^{0}\}$
        \For{$k = K - 1$ \textbf{to} $0$ \textbf{step} $-1$}
            \If{$K-k$ \textbf{is odd}}
                \State \textbf{Direction:} Forward
                \For{$t = 1$ \textbf{to} $T$}
                    \State $y_{t}^{k-1}=f_{\theta}(y_{t}^{k}, y_{t-1}^{k}, \lambda(k), s_t, d_f)$
                \EndFor
            \Else
                \State \textbf{Direction:} Backward
                \For{$t = T$ \textbf{to} $1$ \textbf{step} $-1$}
                    \State $y_{t-1}^{k-1}=f_{\theta}(y_{t-1}^{k}, y_{t}^{k}, \lambda(k), s_{t-1}, d_b)$
                \EndFor
            \EndIf
        \EndFor
    \end{algorithmic}
\end{algorithm}

Although it's possible to reverse the entire sequence (starting in the backward direction and then moving to the forward, followed by backward again), we observed no significant differences in the outcomes between these two cases.

\section{Implementation Details}

\subsection{Single image animation}

Our method is trained at a resolution of 256x256, similar to all other methods. We generates 64x64 animations via the BTU-Net, which are subsequently upscaled to 256x256 using the SR3~\cite{saharia2022image}.We trained both the BTU-Net and SR3 from scratch on the entire training dataset, for 50k and 100k iterations with a batch size of 32, respectively. We set the denoising step to $K=1000$ and the learning rate to 1e-5. During testing, we fine-tune model with test appearance condition for 300 iterations with a learning rate of 1e-5. It should be noted that we employ $K=50$ at test time for expedited generation.

\subsection{Person specific animation}

The training settings for the BTU-Net and SR3~\cite{saharia2022image} are identical to those used in the Single image animation setup, with the exception that both the BTU-Net and SR3~\cite{saharia2022image} are trained for 100 epochs each without fine-tuning.

\subsection{BTU-Net architecture}
\label{appendix:btunet}
\begin{figure}[h]
    \centering
    \includegraphics[width=1\textwidth]{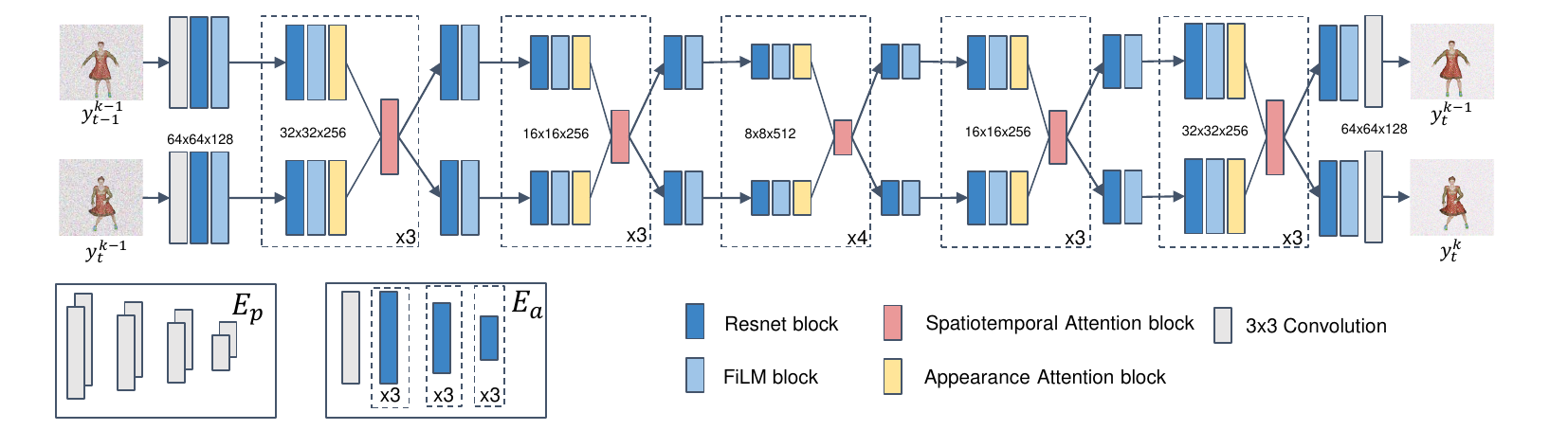}
    \caption{Architecture of BTU-Net.}
    \label{fig:btunet_detail}
\end{figure}

The detailed structural information of the BTU-Net's layers is illustrated in Figure~\ref{fig:btunet_detail}.
Due to the complexity of the arrows, the input directions for $E_a$ and $E_p$ have been omitted. Directions are provided in Figure~\ref{fig:BTUNet} of main manuscript. The notation '$\times digit$' below the dashed block indicates how many times that block structure is repeated.

\subsection{Training and Inference Speed Analysis}
In particular, training the BTU-Net and SR3 models using the UBC fashion dataset requires 15 and 30 epochs, respectively, on a setup of four A100 GPUs, completed within 67 hours. Fine-tuning these models for 300 iterations takes approximately 210 seconds. During inference, processing each frame takes roughly 1.4 to 1.9 seconds on a single A100 GPU.

\section{Ablation Study}

\subsection{Comparison of Bidirectional and Unidirectional Approaches}

\begin{figure}[h]
    \centering
    \includegraphics[width=1\textwidth]{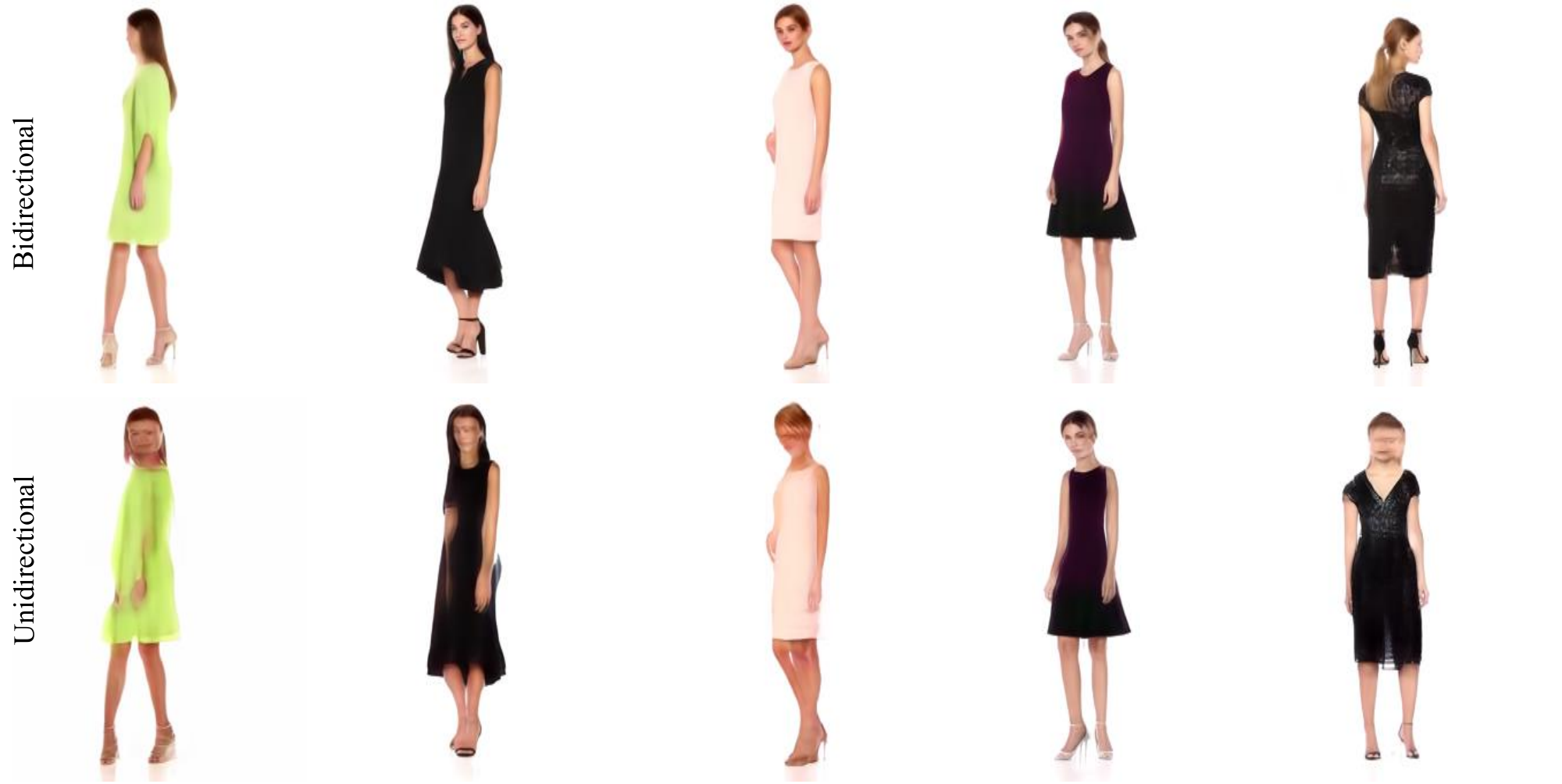}
    \caption{Qualitative comparison of Bidirectional and Unidirectional methods.}
    \label{fig:bi_uni}
\end{figure}

Figure \ref{fig:bi_uni} demonstrates the qualitative results of bidirectional and unidirectional temporal training via our BTU-Net. The unidirectional approach struggles to generate images fitting the pose condition, tending to replicate the texture of the front image input as the condition instead. Unlike the unidirectional approach, the bidirectional model successfully creates images that meet the pose condition.

\begin{figure}[h]
    \centering
    \includegraphics[width=1\textwidth]{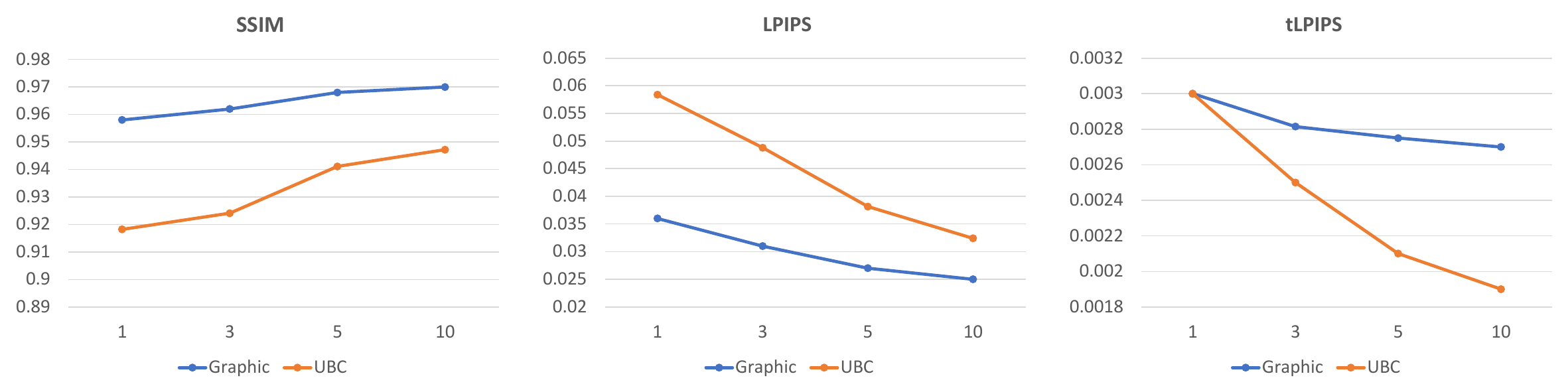}
    \caption{Quantitative comparison of the number of images used for fine-tuning. "Graphic" and "UBC" notates graphic simulated data and UBC fashion data results respectively.}
    \label{fig:num_img}
\end{figure}

\subsection{Number of Images for Fine-Tuning}

\begin{figure}[h]
    \centering
    \includegraphics[width=1\textwidth]{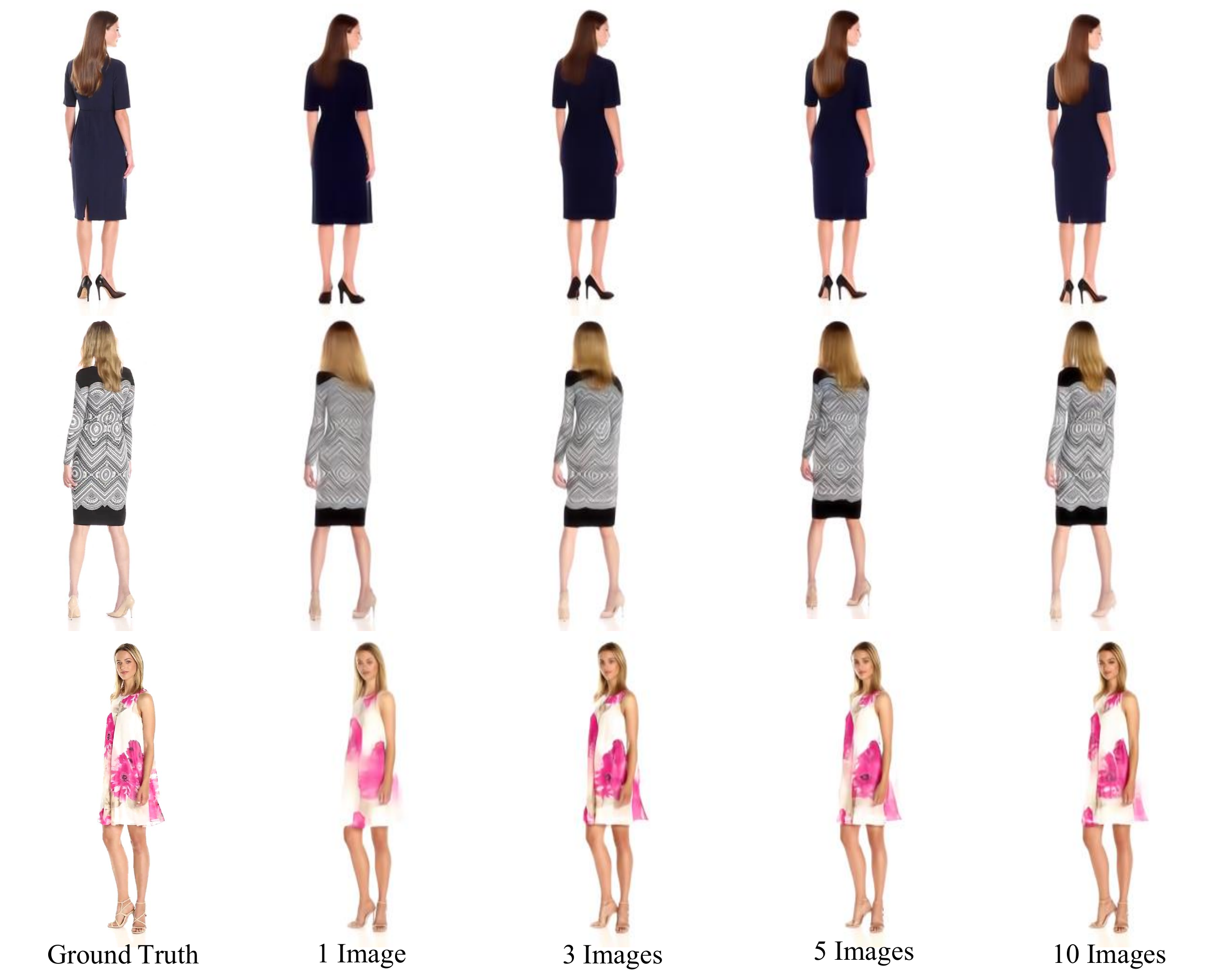}
    \caption{Qualitative comparison of the number of images used for fine-tuning.}
    \label{fig:ft_qual}
\end{figure}

We also evaluate the performance depending on the number of images used for fine-tuning. The performance comparison results are shown in Figure \ref{fig:num_img} and Figure \ref{fig:ft_qual}. As the number of images used for fine-tuning increases, performance improves across various metrics. Notably, the trend between the increase in image count and metric scores is not linear but shows signs of convergence.

\subsection{User Study}
\begin{figure}[h]
    \centering
    \includegraphics[width=1\textwidth]{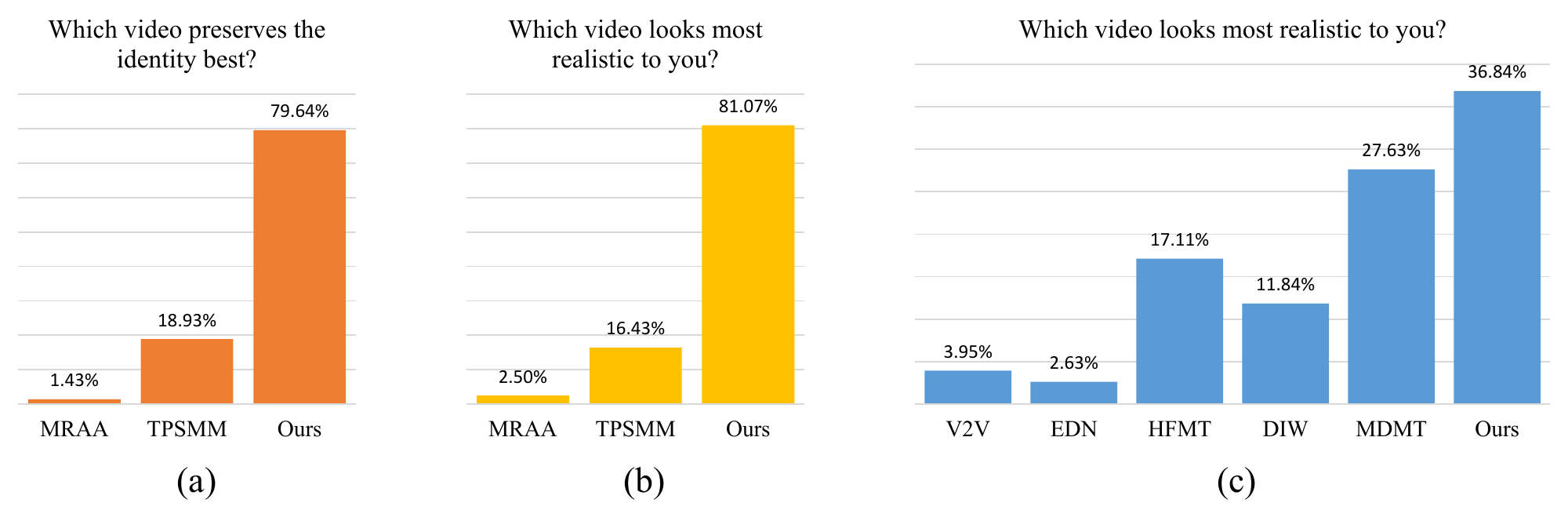}
    \caption{
    Quantitative result of human evaluations. Graphs (a) and (b) represent the results for the \textit{single image animation} task, showing the proportion of choices made for two different questions. Graph (c) shows the results for the \textit{person-specific animation} task.}
    \label{fig:user_quant}
\end{figure}

We conduct quantitative results of an user study in which people evaluated videos generated by our method and baseline methods. A total of 42 participants took part in this study, which involved tasks for \textit{single image animation} and \textit{person specific animation}. Each evaluation required participants to watch comparison videos at least twice and make selections based on two questions: “\textit{Which video preserves the identity best?}” and “\textit{Which video looks most realistic to you?}”. For the person specific animation task, the experiment was conducted excluding the first question.
As shown in Figure \ref{fig:user_quant}, it can be seen that our BTMD results are much more realistic and maintain identity better when evaluated by people, compared to other baseline methods.

\subsection{Unconditional Animation Generation}

\begin{figure}[h]
    \centering
    \includegraphics[width=1\textwidth]{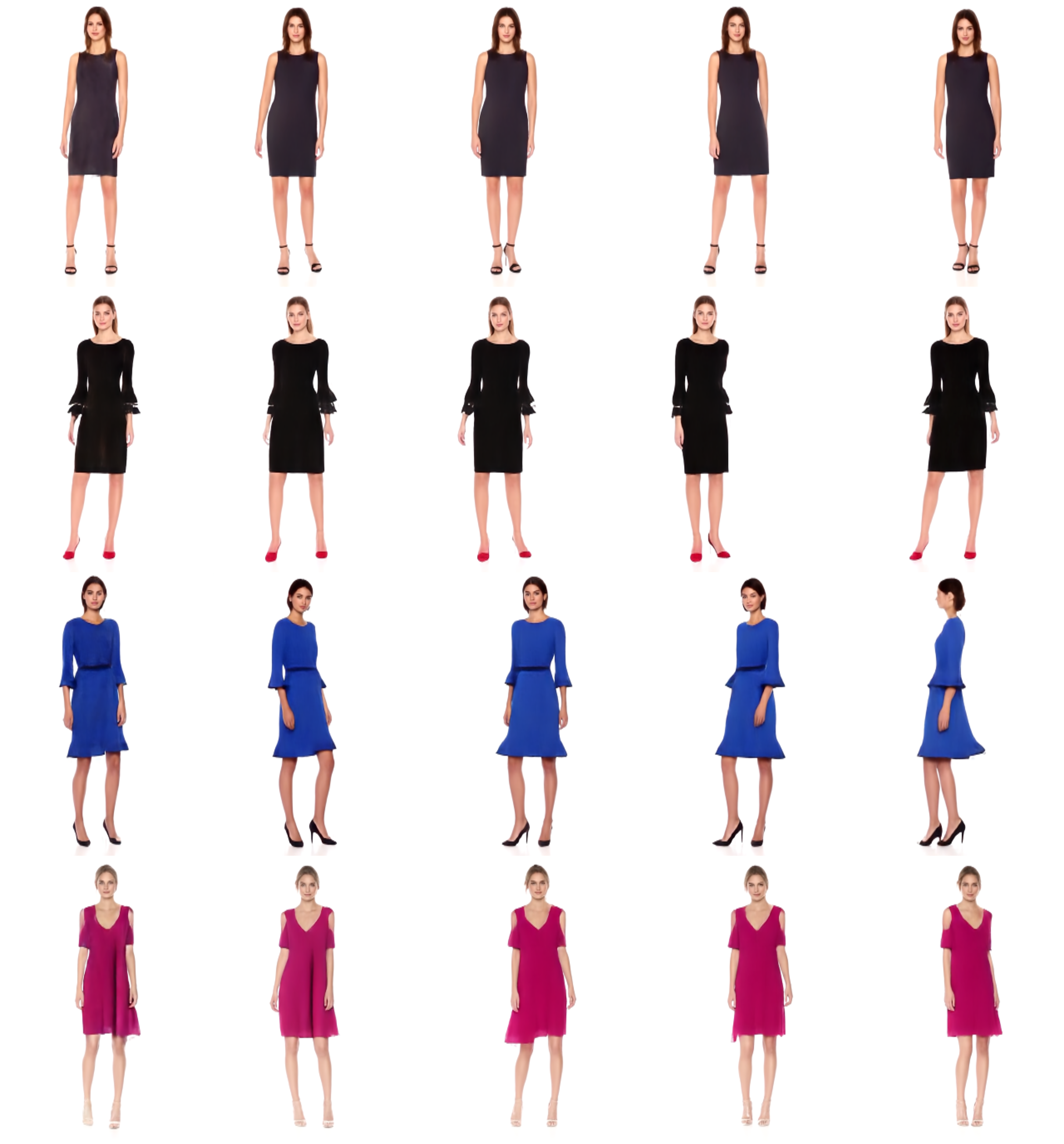}
    \caption{Sample sequences from unconditional generation on UBC fashion data.}
    \label{fig:uncond_ubc}
\end{figure}

\begin{figure}[t]
    \centering
    \includegraphics[width=1\textwidth]{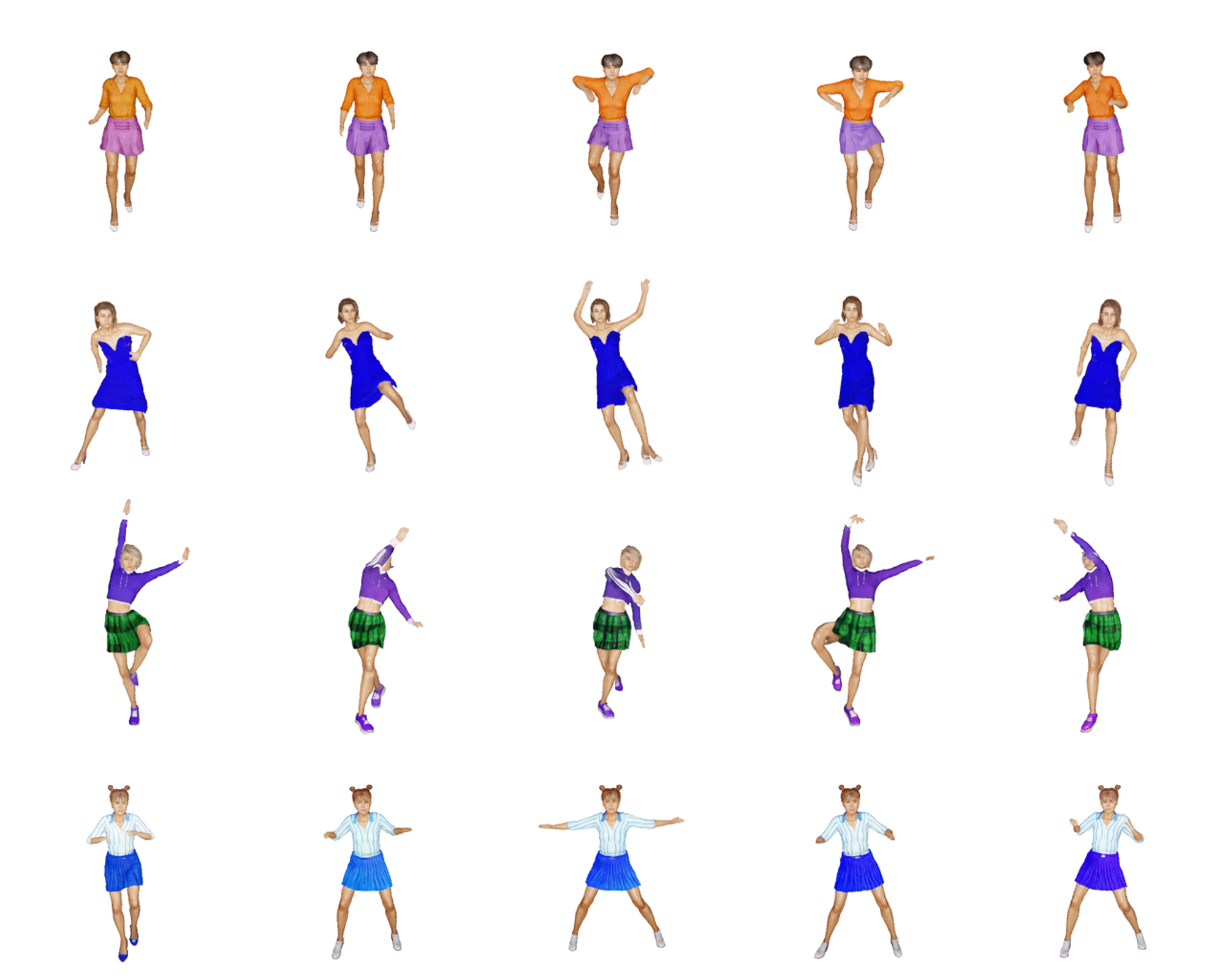}
    \caption{Samples from unconditional generation on Graphic simulated data.}
    \label{fig:uncond_gs}
\end{figure}

We demonstrate that our method can generate human animations featuring diverse clothing styles and identities, even without any image conditions. Results from unconditional generation experiments on both datasets are illustrated in Figures \ref{fig:uncond_ubc} and \ref{fig:uncond_gs}.

\section{Graphic simulated dataset}
\label{appendix:gsdata}

\begin{figure}[h]
    \centering
    \includegraphics[width=1\textwidth]{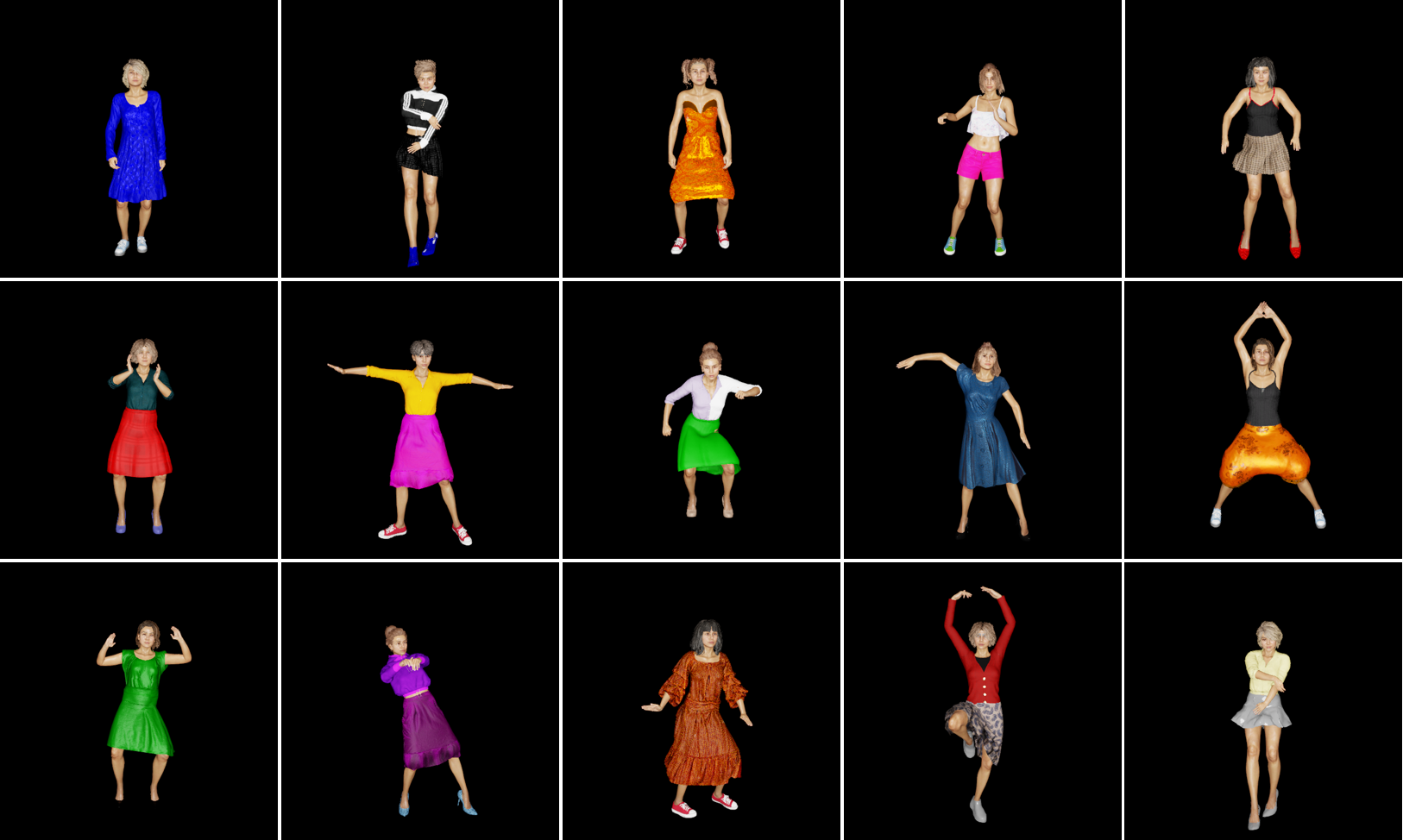}
    \caption{Samples of Graphic Simulation Dataset}
    \label{fig:data}
\end{figure}

Graphic simulated dataset is comprised of approximately 98,000 images, each rendered at a resolution of 512x512. These images illustrate various dynamic movements (such as dance, exercise, etc.) of 3D human models with a total of 99 different appearances. The 3D human models in this dataset are created using Character Creator 4~\cite{chacrea}, and we simulate the soft cloth motion in iClone8~\cite{iClone8}. Mixamo~\cite{mixamo} human motions, are exported as an Alembic~\cite{alembic} file. For realistic rendering, we employ Ray Tracing Texel (RTX) rendering and the Nvidia Omniverse~\cite{omniverse} as the rendering tool. Figure \ref{fig:data} shows a few samples from our graphically simulated data.

\section{More visual results}
\subsection{Single Image animation}
\begin{figure}[h]
    \centering
    \includegraphics[width=1\textwidth]{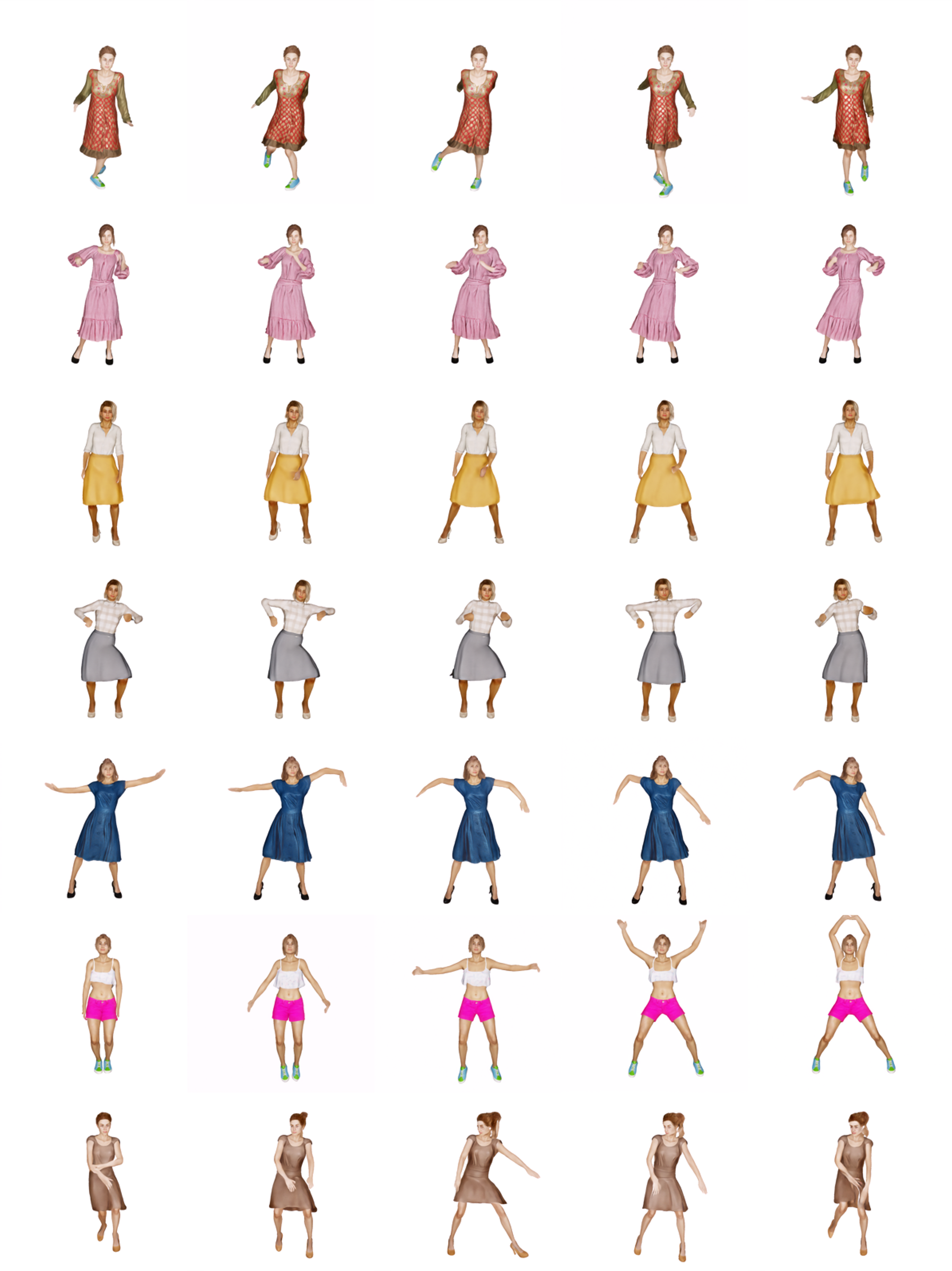}
    \caption{
    More single image animation results on Graphic simulated data.
    }
    \label{fig:gs_more}
\end{figure}

\begin{figure}[h]
    \centering
    \includegraphics[width=1\textwidth]{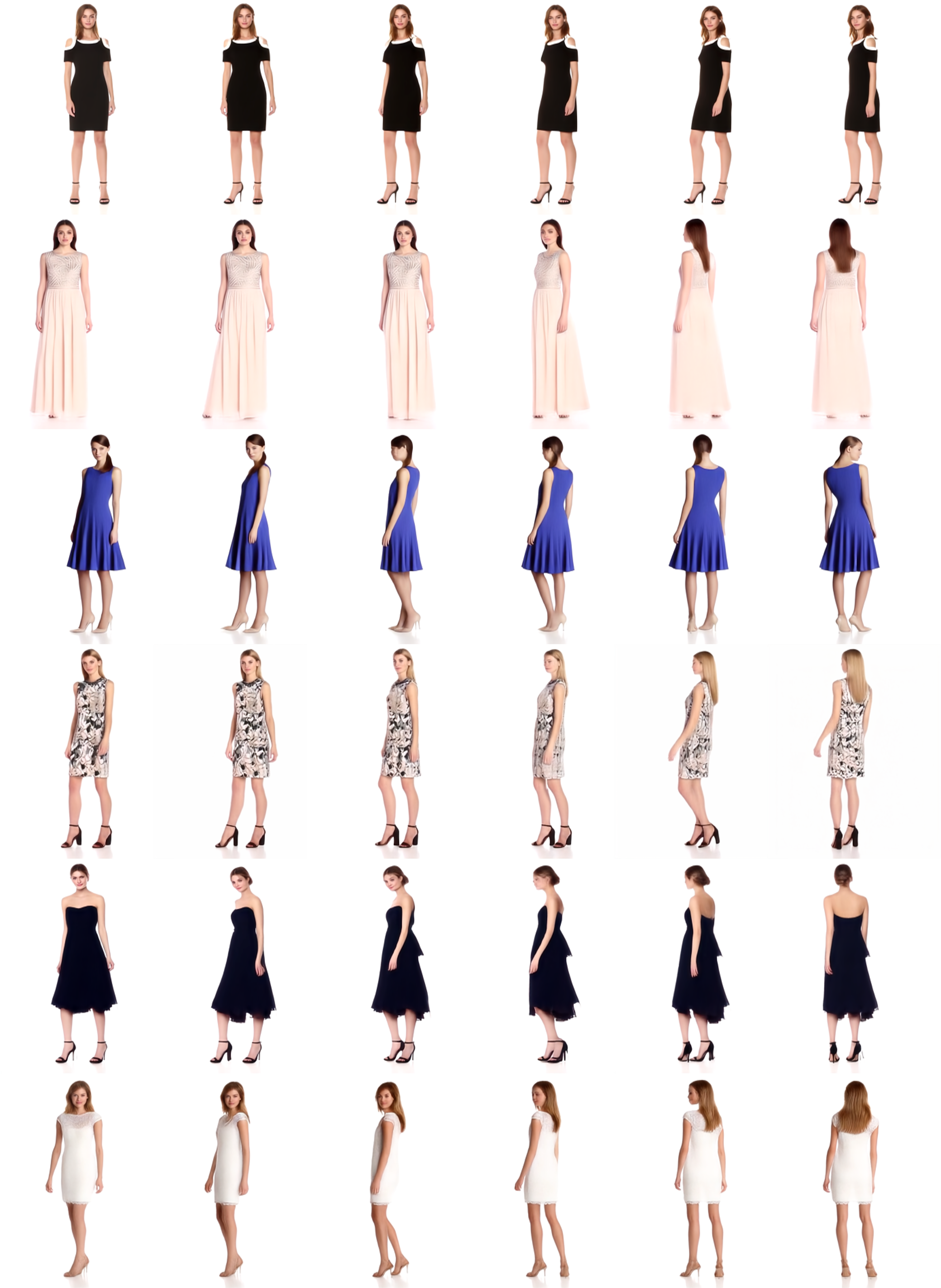}
    \caption{
    More single image animation results on UBC fashion data.
    }
    \label{fig:ubc_more}
\end{figure}

Additional visual results conducted with graphic simulated and UBC fashion data for the single image animation task are shown in Figures \ref{fig:gs_more} and \ref{fig:ubc_more}.

\subsection{Person-specific animation}
\begin{figure}[t]
    \centering
    \includegraphics[width=1\textwidth]{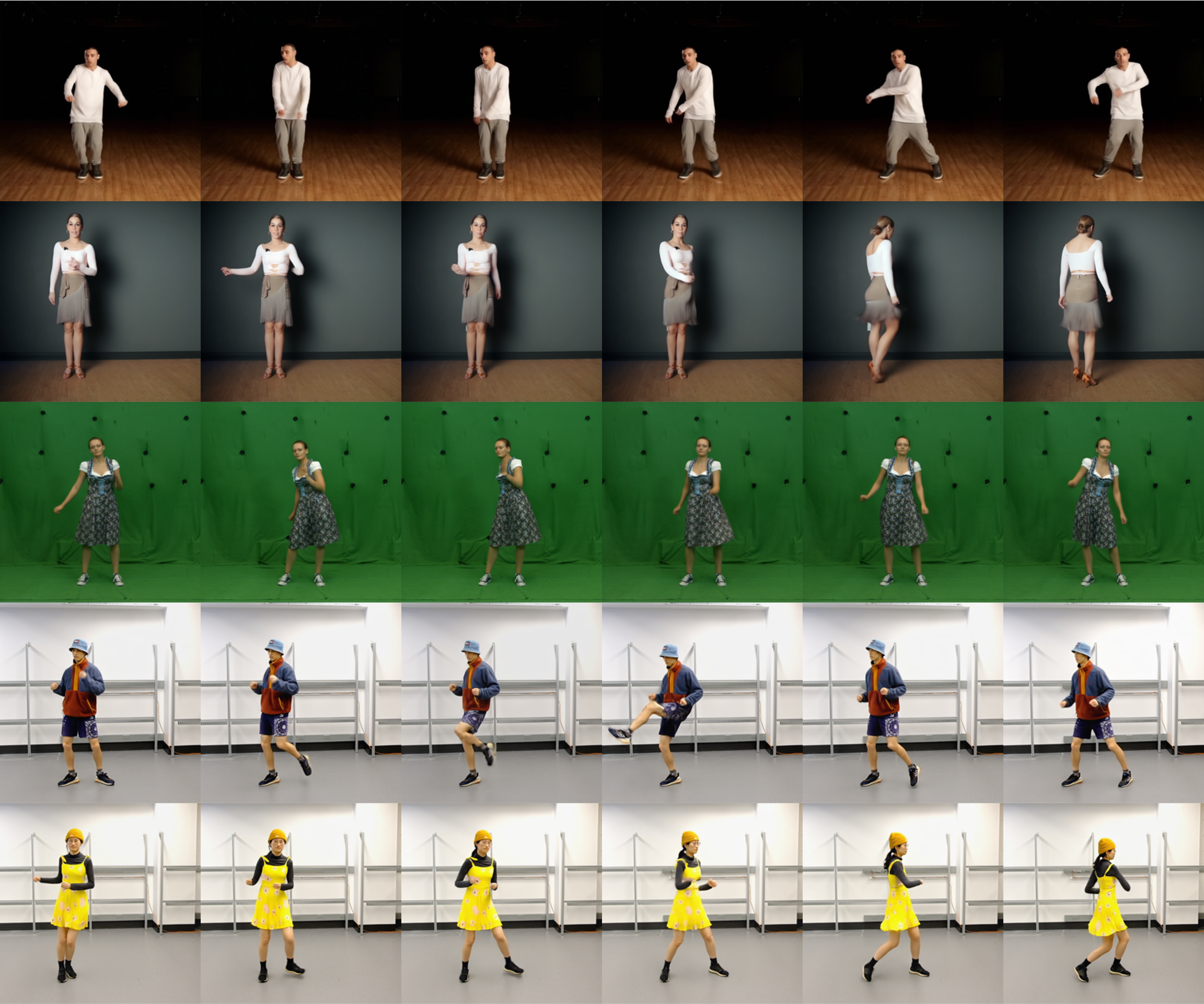}
    \caption{
    Samples from person-specific animation results with background.
    }
    \label{fig:bg_synt}
\end{figure}

For a fair comparison evaluation in the person-specific human animation task, we evaluate the results using only the foreground. However, our method is capable of background synthesis, and the visual results are shown in Figure \ref{fig:bg_synt}.

\end{document}